\documentclass{article}

\usepackage[round]{natbib}

\usepackage[titletoc]{appendix}
\usepackage{amsmath,amsbsy,amssymb,amsfonts,amsthm}
\usepackage{mathtools}
\usepackage{color}
\usepackage{xspace} 
\usepackage{times}
\usepackage{graphicx,subfigure}
\usepackage{algorithm,algorithmic} 

\usepackage[utf8]{inputenc} % allow utf-8 input
\usepackage[T1]{fontenc}    % use 8-bit T1 fonts
\definecolor{mydarkblue}{rgb}{0,0.08,0.45}
\usepackage[colorlinks=true,
    linkcolor=mydarkblue,
    citecolor=mydarkblue,
    filecolor=mydarkblue,
    urlcolor=mydarkblue]{hyperref}      
\usepackage{url}            
\usepackage{booktabs,tabularx}     
\usepackage{nicefrac}      
\usepackage{microtype}     
\newcolumntype{Y}{>{\raggedright\arraybackslash}X}

\newcommand{\norm}[1]{\left\Vert#1\right\Vert}
\newcommand{\lrpar}[1]{\left( #1 \right)}
\newcommand{\absval}[1]{\lvert#1\rvert}

\DeclareMathOperator*{\argmax}{arg\,max}
\DeclareMathOperator*{\argmin}{arg\,min}

\newcommand{\EV}{\mathbb{E}}

\renewcommand{\vec}[1]{\ensuremath{\boldsymbol{#1}}}

\newcommand{\mat}[1]{\ensuremath{\boldsymbol{#1}}}

\newcommand{\bb}{{\mathbf b}}

\newcommand{\bu}{{\mathbf u}}
\newcommand{\bv}{\ensuremath{\boldsymbol{v}}}
\newcommand{\bw}{{\mathbf w}}
\newcommand{\bx}{\ensuremath{\boldsymbol{x}}}
\newcommand{\by}{\ensuremath{\boldsymbol{y}}}

\newcommand{\cA}{\mathcal{A}}
\newcommand{\cL}{\mathcal{L}}
\newcommand{\cQ}{\mathcal{Q}}
\newcommand{\reals}{\mathbb{R}}

\newtheorem{thm}{Theorem}
\newtheorem{prop}[thm]{Proposition}
\newtheorem{lemma}[thm]{Lemma}
\newtheorem{cor}[thm]{Corollary}
\newtheorem{definition}[thm]{Definition}

\usepackage[accepted]{icml2020}

\icmltitlerunning{Linear Lower Bounds and Conditioning of Differentiable Games}

\begin{document}

\twocolumn[
\icmltitle{Linear Lower Bounds and Conditioning of Differentiable Games}




\begin{icmlauthorlist}

\icmlauthor{Adam Ibrahim}{mila}
\icmlauthor{Wa\"iss Azizian}{ens}
\icmlauthor{Gauthier Gidel}{mila}
\icmlauthor{Ioannis Mitliagkas}{mila}
\end{icmlauthorlist}

\icmlaffiliation{mila}{Mila, University of Montreal}
\icmlaffiliation{ens}{Ecole Normale Supérieure, Paris}

\icmlcorrespondingauthor{Adam Ibrahim}{<first>.<last>@umontreal.ca}

\icmlkeywords{Machine Learning, ICML}

\vskip 0.3in
]



\printAffiliationsAndNotice{\icmlEqualContribution} % otherwise use the standard text.

\begin{abstract}
Recent successes of game-theoretic formulations in ML have caused a resurgence of research interest in differentiable games. Overwhelmingly, that research focuses on methods and upper bounds on their speed of convergence. In this work, we approach the question of fundamental iteration complexity by providing lower bounds to complement the linear (i.e. geometric) upper bounds observed in the literature on a wide class of problems. We cast saddle-point and min-max problems as 2-player games. We leverage tools from single-objective convex optimisation to propose new linear lower bounds for convex-concave games. Notably, we give a linear lower bound for $n$-player differentiable games, by using the spectral properties of the update operator. We then propose a new definition of the condition number arising from our lower bound analysis. Unlike past definitions, our condition number captures the fact that linear rates are possible in games, even in the absence of strong convexity or strong concavity in the variables.
\end{abstract}

\section{Introduction}
Game formulations arise commonly in many fields, such as game theory \citep{harker1990finite}, machine learning \citep{kim2008minimax,goodfellow2014generative}, and computer vision \citep{chambolle2011first,wang2014multi} among others,
and encompass saddle-point problems~\citep{palaniappan2016stochastic,chambolle2011first,chen2017accelerated}.
% Notably, training GANs can be formulated as a two-player game where the goal is for the generator and the discriminator to minimise their respective objective, which may be subject to adversarial constraints (e.g. zero-sum) \citep{salimans2016improved,mescheder2017numerics}.
    
The machine learning community has been overwhelmingly using gradient-based methods to train differentiable games~\citep{goodfellow2014generative,salimans2016improved}.
These methods are not designed with game dynamics in mind \citep{mescheder2017numerics}, and to make matters worse, have been often tuned suboptimally \citep{gidel2019negative}.
A recent series of publications in machine learning brings in tools from the minimax and game theory literature to offer better, faster alternatives~\citep{daskalakis2018training, gidel2018variational, gidel2019negative}.
This exciting trend begs the question: how fast can we go? Knowing the fundamental limits of this class of problems is critical in steering future algorithmic research. 
    
In order to answer this question, the optimisation literature contains a few different approaches based on the distance between the iterates at a step $t$ and the optimal choice of parameters. Given an optimisation algorithm, it is possible to show under certain assumptions on the objectives that this error is in $\mathcal{O}\lrpar{\rho^t}$, where the rate of convergence $\rho$ depends on the algorithm \citep{nesterov2013introductory}. If $\rho\in(0, 1)$, we say that the rate of convergence is linear, which corresponds to the error decaying exponentially fast. Hence, a lower bound on the rate of convergence limits how fast an algorithm may converge. This is important as it helps establish the tightness of upper bounds, which happens when they are matched by the lower bounds, and may otherwise indicate possible acceleration of the method considered. For example, in single-objective optimisation, the lower bound on the rate of convergence of first-order black box algorithms is known to be linear for smooth, strongly convex objectives, and can be derived via a domino-like coverage argument by~\citet{nesterov2013introductory}. Another recent, spectral approach by~\citet{arjevani2016lower} complements these results by proposing linear lower bounds for a large class of optimisers in finite-dimensional settings. As the lower bounds for Nesterov's accelerated gradient obtained by those techniques match the upper bound, we know that Nesterov's accelerated gradient is optimal within a large class of methods for smooth, strongly convex objectives. Additionally, in optimisation, a natural concept of condition number arises to describe the difficulty of $\mu$-strongly convex, $L$-smooth objectives. This condition number is the only problem-dependent quantity that appears in both the upper and lower bounds and is given by $\kappa = L / \mu$ \ \citep{nesterov2013introductory}.
In single-objective optimisation, there is a clear distinction between strongly convex objectives, where the condition number is finite and linear rates are achievable, and general convex objectives where the condition number can be undefined and only sublinear rates are possible in general.
    
When studying lower bounds for convex-concave min-max, one is faced with a number of distinct challenges compared to the optimisation setting. In particular, there is no universally accepted definition of a condition number. Some commonly used definitions, like the one used in~\citet{chambolle2011first,palaniappan2016stochastic}, are undefined for bilinear problems, which lack strong convexity and strong concavity in the variables. This is problematic because we know that both extragradient and gradient methods with negative momentum achieve linear convergence in bilinear games \citep{korpelevich1976extragradient, gidel2019negative}. \emph{Can we get a condition number that captures the fact that linear rates are possible even in the absence of strong convexity and strong concavity?}

We show that it is possible by providing new lower bounds, obtained by casting saddle-point and min-max problems as games and leveraging existing proof techniques originally designed for smooth strongly convex, single-objective optimisation. These bounds also yield a meaningful condition number for the bilinear case, in the absence of strong convexity and strong concavity in the variables. Our contributions are summarised as follows:
\begin{enumerate}
    \item We generalise \emph{Nesterov's domino argument} and design a difficult min-max problem to derive a linear lower bound on the rate of convergence of several first-order black box optimisation algorithms for 2-player games and min-max problems.\label{rq:1}
    In order to get an asymptotic rate using the domino bound, one needs to resort to the analysis of infinite-dimensional problems.
    \item We propose a linear lower bound for finite-dimensional problems by generalising the $p$-SCLI framework proposed by \citet{arjevani2016lower} to $n$-objective optimisation algorithms. This lower bound stems from the spectral properties of the algorithms on quadratics, and is valid for any number of players, and in particular 2-player games and min-max problems. This bound is tight for $n=1$ since it reduces to the one presented by \citet{arjevani2016lower} for strongly convex, smooth single-player optimisation. \label{rq:2}
    \item We provide a formulation of the condition number of 2-player games consistent with the existing literature on upper bounds for games and min-max problems. In particular, this condition number is finite for bilinear games. \label{rq:3}
\end{enumerate}

After the results of this work were made available online, several researchers have proposed methods to match some of our bounds in the smooth strongly-convex-strongly-concave setting \citep{fallah2020optimal,lin2020near} and the bilinear setting \citep{azizian2020accelerating}, which is merely convex-concave, thereby establishing the tightness of some of the bounds and the optimality of those methods in those regimes.

The rest of the paper is organised as follows. We purposely discuss preliminaries first in Section~\ref{sec:preliminaries} to introduce the general framework used to present in Section~\ref{sec:background} the relevant literature in the context of our results. In Section~\ref{sec:lower-domino}, we provide lower bounds 
using Nesterov's domino argument, and in Section~\ref{sec:lower-spectral} we improve on those bounds using the spectral technique. We conclude with some discussion.

\section{Preliminaries}
\label{sec:preliminaries}
\subsection{Differentiable games}
Following the definition of \citet{balduzzi2018mechanics}, a \textit{differentiable game} is characterised by $n$ players, each associated with a set of parameters $\bw_i \in \reals^{d_i}$ and a twice continuously differentiable objective function $l_i:\reals^d \rightarrow \reals$ of all the parameters $\bw = (\bw_1,...,\bw_n) \in \reals^d$, where $d = \sum_{i=1}^n d_i$. In particular, if $\sum_{i=1}^n l_i(\bw) = 0$, we say that the game is \textit{zero-sum}. 

Often, we seek to minimise the objectives $l_i$, and look for \textit{Nash equilibria} $\bw^* = (\bw_1^*,...,\bw_n^*)$, which satisfy\footnote{Of course, we could be trying to maximise some players' objectives, but we can without loss of generality work with minima since $\argmax f = \argmin (-f)$} for all $i$
\begin{align}\label{eq:game-Nasheq}
    \bw_i^* \in \argmin_{\bw_i} l_i\lrpar{\bw_1^*, ..., \bw_{i-1}^*, \bw_i, \bw_{i+1}^*, ... , \bw_n^*}\,.
\end{align}
In order to find the Nash equilibria, we may look for stationary points, corresponding to the zeros of the vector field $\bv(\bw) = \lrpar{\nabla_{\bw_1}l_1(\bw) \dots \nabla_{\bw_n}l_n(\bw)}^\top$. In single-objective optimisation, which corresponds to a 1-player game, we know that stationary points of $\bv$ do not necessarily represent minima of the objective function, and higher order information, such as the Hessian, is necessary to determine whether a stationary point is a minimum. The same is true for a game with several players \citep{balduzzi2018mechanics}, where the \emph{Jacobian} of $\bv$, given by
\begin{align} \label{eq:jacobian}
    \nabla \bv(\bw) = \begin{pmatrix}
\nabla^2_{\bw_1} l_1(\bw) & \dots & \nabla_{\bw_n}\nabla_{\bw_1} l_1(\bw)\\
 \vdots & & \vdots \\ 
 \nabla_{\bw_1}\nabla_{\bw_n} l_n(\bw) &\dots & \nabla^2_{\bw_n} l_n(\bw)
\end{pmatrix}
\end{align}
gives sufficient conditions to determine whether a stationary point is a Nash equilibrium. Note that our lower bound analysis encompasses games with stable stationary points that are not Nash equilibria. 

\subsection{Quadratic games}
In order to gain insight on general games, we focus on quadratic games\footnote{Note that quadratic games are inherently relevant; e.g. in reinforcement learning to learn a linear value function from the mean squared projected Bellman error \citep{du2017stochastic}}, corresponding to games with quadratic objectives $l_i$. In our analysis, we will mostly discuss two-player games, where the players respectively control the parameters $\bx \in \reals^{d_1}$ and $\by \in \reals^{d_2}$. The quadratic objectives take the form
\begin{align}\label{eq:quad-vec-field-2-losses}
    l_1(\bx, \by) &= \frac{1}{2} \bx^\top \mat{S}_1 \bx + \bx^\top \mat{M}_{12} \by + \bx^\top \bb_1 \nonumber \\
    l_2(\bx, \by) &= \frac{1}{2} \by^\top \mat{S}_2 \by + \by^\top \mat{M}_{21} \bx + \by^\top \bb_2
\end{align}
with $\mat{S}_1$ and $\mat{S}_2$ symmetric. 
In that case the vector field is given by
\begin{alignat}{2}\label{eq:quad-vec-field-2-gen}
    &\bv(\bx, \by) &&= \begin{pmatrix}
    \mat{S}_1 \bx + \mat{M}_{12} \by + \bb_1\\
    \mat{M}_{21} \bx + \mat{S}_2 \by + \bb_2
    \end{pmatrix} \nonumber\\
    & &&= \mat{A} \begin{pmatrix}
    \bx \\
    \by
    \end{pmatrix} + \bb \\ 
    &\text{ with } \mat{A} &&\triangleq \begin{pmatrix}
    \mat{S}_1 & \mat{M}_{12} \\
    \mat{M}_{21} & \mat{S}_2
    \end{pmatrix}, \quad \bb \triangleq \begin{pmatrix}
    \bb_1 \\
    \bb_2
    \end{pmatrix}\nonumber
\end{alignat}
where $\mat{A}$ is the Jacobian of $\bv$. For $n$-player quadratic games, the vector field and Jacobian $\mat{A}$ take the form
\begin{align}\label{eq:quad-vec-field-n}
    &\bv(\bw) = \mat{A}\bw + \bb \\
    &\text{with }\mat{A} \triangleq \begin{pmatrix}
 \mat{S}_{1} & \dots & \mat{M}_{1n}\\
 \vdots & & \vdots \\ 
 \mat{M}_{n1} &\dots & \mat{S}_{n}
\end{pmatrix}, 
\quad \bb \triangleq \begin{pmatrix}
\bb_1 \\
\vdots \\
\bb_n
\end{pmatrix}\nonumber
\end{align}
where the $\mat{S}_i$ are symmetric. For more details on quadratic games, see Appendix \ref{apdx:n-player-games}. Further assumptions on the dimensionality or properties of $\mat{S}_i$ will be introduced in the rest of the paper as they become relevant (e.g. positive semi-definiteness in the next subsection). 

We shall henceforth refer to the Jacobian of the vector field of quadratic $n$-player games simply as the Jacobian, since our analysis will be solely based on quadratic objectives. Interestingly, the problem of finding $\bw$ such that $\bv(\bw) = 0$ consists in solving a system of linear equations (SLE) \citep{richardson1911ix}. In fact, several techniques to precondition systems of linear equation make use of casting the SLE as a game and optimising it with proximal methods, such as \citet{benzi2004preconditioner}.

In this paper, we will denote the spectrum of a matrix $\mat{M}$ by $\sigma(\mat{M})$, and define the \textit{block spectral bounds} $\mu_1, \mu_2, \mu_{12}, L_1, L_2, L_{12}$ as constants bounding the spectra of the blocks in the Jacobian of eq. \ref{eq:quad-vec-field-2-gen}:
\begin{align}\label{eq:def_spectrum_bounds}
    \mu_1 \leq \vert \sigma(&\mat{S}_1) \vert \leq L_1 \quad\quad
    \mu_2 \leq \vert \sigma(\mat{S}_2) \vert \leq L_2 \quad\quad \nonumber \\
    &\mu_{12}^2 \leq \vert \sigma(\mat{M}_{12} \mat{M}_{12}^\top) \vert \leq L_{12}^2
\end{align}
where we assume that $\mat{M}_{12}$ is a wide or square matrix (if it is a tall matrix, we use $\mu_{12}^2 \leq \vert \sigma(\mat{M}_{12}^\top \mat{M}_{12}) \vert \leq L_{12}^2$ to define $\mu_{12}$ and $L_{12}$ instead of the last inequality of eq. \ref{eq:def_spectrum_bounds}).

\subsection{Min-max of quadratics as 2-player quadratic games}
Consider the family $\mathcal{P}$ of min-max problems of the form 
\begin{alignat}{2}\label{eq:minmax}
    \min_{\bx\in \mathbb R^{d_1}} \max_{\by \in \mathbb R^{d_2}} &\quad f(\bx, \by) = \bx^\top \mat{M} \by + \frac{1}{2} \bx^\top \mat{S}_1 \bx - \frac{1}{2} \by^\top \mat{S}_2 \by \nonumber\\
    &\qquad\qquad\qquad+ \bx^\top \bb_1 - \by^\top \bb_2 + c \tag{$\mathcal{P}$} \\
    \text{ where} &\quad \sigma(\mat{MM}^\top),\ \sigma(\mat{S}_1),\  \sigma(\mat{S}_2)\subseteq [0, +\infty) \nonumber
\end{alignat}
with possible constraints and where the dimension need not be finite, e.g. $\bx, \by \in \ell_2 \triangleq \{\bu \in \reals^\mathbf{N} \mid \sum_i^\infty \bu_i^2 < \infty\}$. The optimisation of such a problem is equivalent to finding a pair $(\bx^*,\by^*)$ such that,
\begin{align} \label{eq:prelim_P_nasheq}
\bx^* \in \argmin f(\bx, \by^*) \ \text{ and } \ \by^* \in \argmax f(\bx^*, \by)
\end{align}
%\todo{discuss existence of such point here.}
Noting that $\argmax f = \argmin (-f)$, we get that this optimisation problem is equivalent to a zero-sum 2-player game with objectives $l_1 = -l_2 = f$ (see eq. \ref{eq:game-Nasheq}). This problem can be reduced to searching for the Nash equilibria of the 2-player quadratic game with simplified objectives $l_{\bx}(\bx, \by) = \frac{1}{2} \bx^\top \mat{S}_1 \bx + \bx^\top \mat{M} \by + \bx^\top \bb_1$ and $l_{\by}(\bx, \by) = \frac{1}{2} \by^\top \mat{S}_2 \by - \bx^\top \mat{M} \by + \by^\top \bb_2$, where the $\mat{S}_i$ have been symmetrised (see Appendix \ref{apdx:n-player-games} for an explanation of the symmetrisation of $\mat{S}_i$ and why the objectives can be simplified). Eq. \ref{eq:quad-vec-field-2-gen} yields the vector field
\begin{align}\label{eq:quad-vec-field-2-SPP}
    \bv(\bx, \by) = \begin{pmatrix}
    \mat{S}_1 & \mat{M} \\
    -\mat{M}^\top & \mat{S}_2
    \end{pmatrix}
    \begin{pmatrix}
    \bx \\
    \by
    \end{pmatrix}
    +\begin{pmatrix}
    \bb_1 \\
    \bb_2
    \end{pmatrix}
\end{align}
Therefore, the pair $(\bx^*, \by^*)$ from eq. \ref{eq:prelim_P_nasheq} exists if and only if the corresponding games with vector field given in eq. \ref{eq:quad-vec-field-2-SPP} admit a Nash equilibrium since $\mat{S}_1,\mat{S}_2 \succeq 0$. Note that we could also go from a quadratic game satisfying the above to a min-max formulation. Hence, any lower bound on quadratic games of the form of eq. \ref{eq:quad-vec-field-2-SPP} is a lower bound on min-max problems (in \ref{eq:minmax}), and vice versa.

\section{Background}
\label{sec:background}
\subsection{Existing bounds for 2-player quadratic min-max problems}
Some upper bounds on the rate of convergence of certain optimisation algorithms exist for unconstrained problems in \ref{eq:minmax}. These upper bounds on the rate of convergence $\rho$ imply that for any problem in \ref{eq:minmax}, the iterates will converge to a solution in $\mathcal{O}\lrpar{\rho^t}$. For clarity's sake, we reformulate these upper bounds to be consistent with the notation of \ref{eq:minmax} and eq. \ref{eq:def_spectrum_bounds}. Letting $\kappa = \frac{L_{12}}{\sqrt{\mu_1 \mu_2}}$, \citet{chen1997convergence} analyse the forward-backward algorithm, and find a convergence in $\mathcal{O}\lrpar{\lrpar{\sqrt{1-\lrpar{\frac{\min (\mu_1, \mu_2, \mu_{12})}{\max(L_1, L_2, L_{12})}}^2}}^t}$. \citet{chambolle2011first} give an algorithm for which the convergence is in $\mathcal{O}\lrpar{\lrpar{\sqrt{1 - \frac{2}{\kappa+2}}}^t}$. \citet{palaniappan2016stochastic} present an accelerated version of the forward-backward algorithm with variance reduction with convergence in $\mathcal{O}\lrpar{\lrpar{1 - \frac{1}{1 + 2\kappa}}^t}$. Note that asymptotically, the Chambolle-Pock and accelerated Forward-Backward rates match up to a factor of $2$ on $\kappa$. Finally, \citet{gidel2019negative} give an upper bound in $\mathcal{O}\lrpar{\lrpar{1 - \frac{1}{4 L_{12}^2/\mu_{12}^2}}^t}$ on the convergence of alternating gradient descent with negative momentum for non-singular \emph{bilinear games}, i.e. quadratic games satisfying eq. \ref{eq:quad-vec-field-2-SPP} with $\mat{S}_1 = \mat{S}_2 = 0$ and non-singular Jacobian.

A key problem with those rates of convergence is that the tightness of the upper bound is not established. Such information is important since it may indicate that the algorithm can be accelerated. Ideally, one would use the rate of convergence of the hardest problem (i.e. slowest convergence) in the class of problems, which would be a tight upper bound. If one can only find a lower bound on the rate of convergence of the hardest problem, then any upper bound on the entire problem class must be greater than that lower bound to avoid a contradiction. This is because the (upper bound on the) rate of convergence for a class of problems must apply to \emph{any} problem in the class, and hence be greater than any lower bound derived on any particular problem within that class. Usually, it is not possible to find the problem with the slowest convergence, so one may have to guess a hard enough problem. If the lower bound on that problem matches the upper bound for the problem class, then we have established that not only this problem is one of the hardest problems in the class, but also that the upper bound is tight. Therefore, it is important to remember that the goal of lower bounds generally is not to apply to every problem within the class, unlike upper bounds, but to help estimate how much the upper bounds on the whole class can be improved given the presence of hard problems in the class.

Unlike upper bounds, relevant lower bounds for first-order methods on saddle-point problems are scarcer in the literature. \citet{nemirovsky1992information} gives a lower bound in $\mathcal{O}(1/t)$ for a limited number of steps. \citet{ouyang2018lower} also leverage Krylov subspace techniques, and show lower bounds in $\mathcal{O}(1/t)$ in the monotone case and $\mathcal{O}(1/t^2)$ in the strongly monotone case, assuming the number of iterations is less than half the dimension of the parameters. Note that \citet{ouyang2018lower} do not assume smoothness of the objective in $\by$. A key issue is that since these bounds are only valid for a limited number of steps, they do not yield bounds that can be compared with the upper bounds previously mentioned. In contrast, the lower bounds presented in this work are valid for any number of steps and are linear, and therefore provide a direct limit to the acceleration of methods achieving linear convergence on two-player games. Additionally, our lower bounds also yield condition numbers that give intuition about the difficulty inherent to a problem, and can be computed in a plug-and-play fashion using either bounds on the spectrum of the full Jacobian, or on the spectra of its blocks.

\subsection{Lower bound techniques for convex optimisation with bounded spectrum}
In single-objective optimisation, i.e. a 1-player game, the Jacobian in eq. \ref{eq:jacobian} reduces to the Hessian of the objective, denoted $\mat{H}(\bx)$. In that case, if there exists $\mu, L \in \reals^{++}$ such that for all $\bx$ in the domain considered $$\mu \preceq \mat{H}(\bx) \preceq L$$ the objective is $\mu$-strongly convex and has $L$-Lipschitz gradients, and the convergence rates (i.e. upper bounds) are known to be linear in the number of iterations for several classes of algorithms~\citep{nemirovsky1983problem,nesterov2013introductory}. In the context of convex minimisation, various lower bounds have been derived depending on whether the objective is strongly convex and/or has Lipschitz gradients (see \citep{bubeck2015convex} for an overview). 

\textbf{Nesterov's lower bound} In particular, \citet{nesterov2013introductory} gives an information-based complexity bound for $\mu$-strongly convex objectives with $L$-Lipschitz gradients, by showing that there is a $\mu$-strongly convex example in $\ell_2 \rightarrow \reals$ with $L$-Lipschitz gradients for which first-order black box methods, i.e. methods using only past iterates and gradients of past iterates at every update, converge linearly at a rate at least $\rho = 1 - \frac{2}{\sqrt{\kappa}+1}$, where the condition number is given by $\kappa = L/\mu$. The proof relies on the fact that at iteration $t$, only the $t$ first components of the estimates $\bx_t$ have been updated from their initial values, where $\bx_0 = 0$. This is then used to lower bound the distance to the optimum. Since an infinite number of iterations is required to converge in $\ell_2$ if the solution $\bx^*$ has an infinite number of nonzero components, we obtain asymptotic rates. An important caveat is that an infinite-dimensional example does not directly yield a lower bound for finite-dimensional problems.

\textbf{$p$-SCLI} \citet{arjevani2016lower} introduce the $p$-SCLI framework to provide bounds for a large class of methods used for optimising $\mu$-strongly convex objectives with $L$-Lipschitz gradients. Roughly speaking, an algorithm is $p$-SCLI if its update rule on quadratics $f(\bx) = \bx^\top \mat{A} \bx + \bx^\top \bb$ with $\mat{A}$ symmetric is a linear combination of the $p$ previous iterates and $\bb$, where the coefficients are matrices that depend on $\mat{A}$ and are assumed to be simultaneously triangularisable. The spectral properties of the update rule are used to derive lower bounds on the rate of convergence of $p$-SCLI algorithms. The lower bound on the rate of convergence of $p$-SCLI methods is given by $\rho = 1 - \frac{2}{\sqrt[p]{\kappa}+1}$ for $\kappa = L/\mu$. This allows us to recover lower bounds for gradient descent ($p = 1$) or \citet{nesterov1983method}'s accelerated gradient descent ($p = 2$) that match the upper bounds. A key advantage is that these bounds are more refined than Nesterov's by introducing the dependence on $p$ (for example, both GD and Nesterov's accelerated gradient descent are black box first-order methods, but this bound is tighter for $p=1$ methods such as GD), and do not rely on an infinite-dimensional example, but rather on the spectral properties of the methods. Such bounds highlight how lower bounds can suggest potential acceleration: in the example previously given, the addition of momentum to gradient descent turns the method from $p=1$ to $p=2$, thereby decreasing the lower bound and allowing for potentially faster convergence. Finally, the $p$-SCLI framework also yields upper bounds, and the authors also show a general method to accelerate algorithms on quadratics with the hope that the acceleration is relevant to more general classes of objectives, albeit at a cost too prohibitive to be practical.

Interestingly, both techniques produce a tight lower bound from a quadratic objective, indicating that quadratics are asymptotically as hard as any other strongly convex, smooth problem in single-objective optimisation. This motivates the use of quadratic games to derive the lower bounds presented in this paper as we generalise those methods to the multi-objective setting. When there are several players, however, the Jacobian is no longer symmetric, and its spectrum will generally be complex, and hence several of the arguments used in single-objective optimisation fail to apply directly.

\section{Parametric Lower Bounds from Nesterov's Domino Argument}
\label{sec:lower-domino}
In this section, we will only discuss min-max problems. The class $\mathcal{F}$ of counterexamples considered is
\begin{align}\label{eq:domino_class}
    \min_{\bx\in\ell_2} \max_{\by\in\ell_2} \quad f(\bx, \by) =\ &c\bx^\top \mat{M}\by - d_1 \bx^\top \vec{e_1} + d_2 \by^\top \vec{e_1} \nonumber\\
    &\quad+ \frac{\mu_1}{2} \norm{\bx}^2 - \frac{\mu_2}{2} \norm{\by}^2
    \tag{$\mathcal{F}$}
\end{align}
where $\mat{M}$ is an infinite-dimensional bidiagonal matrix i.e. $\forall i, M_{ii} = a_0$ and $M_{i,i+1} = a_1$ with all other entries set to 0, such that $c a_0 a_1 \neq 0$ and $\mu_1, \mu_2 \in \reals^{++}$. Since these problems are in \ref{eq:minmax}, the lower bounds of this section are in particular bounds on the optimisation of min-max problems.
\begin{definition}[Two-step linear span assumption] \label{def:2-step_linear_span}
A first-order black box method for 2-player games satisfies the two-step linear span assumption on \ref{eq:domino_class} if for problems in \ref{eq:domino_class} with Jacobian $\mat{A}$ (cf eq. \ref{eq:quad-vec-field-n}):
\begin{align}
    \bw_t \in \bw_0 + \text{Span} \big(\bw_0, ..., \bw_{t-1}, \mat{A}\bw_0, ..., \mat{A}\bw_{t-1}, \nonumber\\
    \mat{A}^2 \bw_0, ..., \mat{A}^2 \bw_{t-1}, \bb, \mat{A}\bb\big)
\end{align}
\end{definition}
Examples of such methods include simultaneous gradient descent, negative momentum and extragradient. One way to design challenging problems for these methods is to construct problems with a dense solution $(\bx^*,\by^*)$ for which only one new component of the iterates may change from its initial value at every iteration~\citep{nesterov2013introductory}, a phenomenon we will refer to as the \textit{domino argument} (see Appendix \ref{apdx:domino_argument} for some intuition, where we show that the argument also applies to cases where diagonal matrices are used as coefficients in the span, and to alternating implementations of any algorithm satisfying the two-step linear span assumption on \ref{eq:domino_class} thanks to the properties of bidiagonal Toeplitz matrices). 

\subsection{A first lower bound for games with block spectral bounds $\mu_1, \mu_2, L_{12}$}
\begin{prop}[Naive bound]\label{prop:domino_bad_lower_bound_infdim}
For any problem class containing quadratic games, there exists a function $f:\ell_2 \times \ell_2 \rightarrow \mathbb{R}$ corresponding to a problem in \ref{eq:minmax} with block spectral bounds $\mu_1 = L_1, \mu_2 = L_2, L_{12}\in \reals^{++}$ as defined in eq. \ref{eq:def_spectrum_bounds}, that has condition number $\kappa = \frac{L_{12}}{\sqrt{\mu_1\mu_2}}$ such that for any number of iterations $t \geq 1$ and any procedure satisfying the two-step linear span assumption (see def. \ref{def:2-step_linear_span}), the following lower bound holds:
\begin{align}
    \norm{(\bx_t, \by_t)-(\bx^*, \by^*)} \geq \left(1 - \frac{2}{\sqrt{\kappa^2 + 1} + 1}\right)^{t+1}  \nonumber\\ \cdot\norm{(\bx_0, \by_0) - (\bx^*, \by^*)}
\end{align}
\end{prop}
We invite the reader to consult Appendix~\ref{apdx:domino_bad_lower_bound_infdim_proof} for the proof. This lower bound on the distance to a solution also yields a lower bound on the minimum number of steps necessary for all subsequent iterates to be within a target distance --- typically referred to as \emph{iteration complexity}. We may interpret the proposition as being a bound for 2-player games with block spectral bounds $\mu_1$, $\mu_2$, and $L_{12}$ as the bound provided holds for a problem sharing the same block spectral bounds. Similarly, we also get a bound on problems with block spectral bounds $L_1$, $L_2$ and $L_{12}$ by replacing $\mu_i$ by $L_i$ appropriately in $\kappa$. 

We appear to obtain the same condition number as in the upper bound literature. If we assume this bound and condition number to be representative of a finite-dimensional bound as was the case in convex optimisation, we easily see an apparent contradiction from the upper bound on the rate of convergence of alternating gradient descent with negative momentum for bilinear games given by \cite{gidel2019negative}. Indeed, if we let $\mu_1, \mu_2 \rightarrow 0$, the rate of convergence in Prop. \ref{prop:domino_bad_lower_bound_infdim} goes to 1, whereas the upper bound of negative momentum is not affected and may indicate fast convergence. This illustrates how the condition number of the upper bounds is not general enough to be representative of inherent difficulty: it can be shown that for the problem used in the proof of the proposition, $\mu_{12} = 0$. As such, it is not surprising that the bound failed to hold against the upper bound of negative momentum on bilinear games; they were not comparable as by definition bilinear games have $\mu_{12} > 0$. This shows that $\mu_{12}$ encodes critical information that this condition number was not able to capture. Nevertheless, an important point is that the bound itself is correct and represents a problem with slow convergence; it just fails to yield a condition number that accurately captures difficulty as $\mu_{12}$ does not appear. 

However, by refining our proof technique, we can derive a bound which avoids this issue, and yields tighter bounds for games for which we know $\mu_{12}, L_{12}, \mu_1, \mu_2$.

\subsection{Improved lower bound for games with block spectral bounds $\mu_1, \mu_2, \mu_{12}, L_{12}$}
\begin{thm}\label{thm:domino_good_lower_bound_infdim}
For any problem class containing quadratic games, there exists a function $f:\ell_2 \times \ell_2 \rightarrow \mathbb{R}$ corresponding to a problem in \ref{eq:minmax} with block spectral bounds $\mu_1 = L_1, \mu_2 = L_2, L_{12}\in \reals^{++}, \mu_{12} \in \reals^+$ as defined in eq. \ref{eq:def_spectrum_bounds}, that has condition number $\kappa = \sqrt{\frac{L_{12}^2 + \mu_1 \mu_2}{\mu_{12}^2 + \mu_1 \mu_2}}$, such that for any number of iterations $t \geq 1$ and any procedure satisfying the two-step linear span assumption, the following lower bound holds:
\begin{align}
    \norm{(\bx_t, \by_t)-(\bx^*, \by^*)} \geq &\left( 1 - \frac{2}{\kappa + 1}\right)^{t+1} \nonumber \\ 
    &\cdot \norm{(\bx_0, \by_0) - (\bx^*, \by^*)}
\end{align}
The same result holds for any problem class containing bilinear games, if one sets $\mu_1 = L_1 = \mu_2 = L_2 = 0$.
\end{thm}

\begin{cor}[Iteration complexity bound]
For the same problem classes and under the same assumptions as Theorem \ref{thm:domino_good_lower_bound_infdim}, the minimal number of steps $t$ required to reach a target distance $\epsilon$ from the solution, that is $\norm{(\bx_t, \by_t)-(\bx^*, \by^*)} < \epsilon$, is given by
\begin{align}
    t \geq \frac{\kappa-1}{2}\log\lrpar{\frac{\norm{(\bx_0, \by_0) - (\bx^*, \by^*)}}{\epsilon}} - 1
\end{align}
\end{cor}

This generalises Prop. \ref{prop:domino_bad_lower_bound_infdim}. The proof can be found in Appendix \ref{apdx:domino_good_lower_bound_infdim_proof}, where we also show how we used spectral properties of Toeplitz matrices in Banach algebras to create the hard problems yielding the bound. The proof of the iteration complexity bound can be found in Appendix \ref{apdx:iteration-complexity}. As with the lower bounds, note that the iteration complexity bound does not necessarily hold for every problem in the class; the idea is that to reach a target error $\epsilon$ we know that there is at least one problem that requires at least the number of steps indicated in the bound, and therefore that to optimise over a problem class that satisfies the assumptions, we will need in general at least the number of steps given in the bound, to account for the hard problems. %This bound is very close in its dependence on $\kappa$ to the upper bounds on strongly-convex-strongly-concave problems from \citet{chambolle2011first} and \citet{palaniappan2016stochastic}, albeit our $\kappa$ is more general than theirs. For bilinear games, where $\mu_1=L_1=\mu_2=L_2=0$, the dependence of the lower bound on $\kappa$ differs by a square root from that of the upper bound given by \citet{gidel2019negative}.

It is important to emphasize that as is the case with Nesterov's argument for single-objective optimisation, this bound is still based on an infinite-dimensional problem, and that upper bounds generally are proven for finite-dimensional settings. We may hope that this bound also holds in finite dimension, since we are not aware of upper bounds contradicting it, and were not able to generate finite-dimensional 2-player games for which the bound did not hold empirically. The condition number appearing in Thm. \ref{thm:domino_good_lower_bound_infdim} is more expressive than the one found in the upper bound literature $\kappa = L_{12}/\sqrt{\mu_1\mu_2}$, and is lower bounded by 1, instead of 0. A limitation, however, is that this $\kappa$ is not able to dissociate $L_1 \neq \mu_1, L_2 \neq \mu_2$, which is a problem in terms of expressivity of the condition number. It may also threaten the tightness of the lower bound since intuition from convex optimisation would suggest that objectives with matching lower and upper bounds on the spectra are easier to optimise. This stems from the fact that the closed form solution for problems in \ref{eq:minmax} when $\mat{S}_i$ is non-scalar, which we would need for all block spectral bounds to appear in $\kappa$, is complicated and the associated condition number is impractical. Therefore, we leave the matter of deriving a practical bound based on the domino argument involving all $\mu_i$ and $L_i$ as future work.

Interestingly, the rate takes the same form as in strongly convex smooth optimisation, suggesting that for general $n$-player games, we may still get a lower bound of the form $\rho \geq 1 - \frac{2}{\kappa + 1}$ for some generalised condition number $\kappa$. This intuition will be highlighted in the next section, by deriving lower bounds from the spectral properties of the update operators of a large class of optimisation methods for $n$-player games. The results we are about to introduce will also address the matter of $L_i \neq \mu_i$, and will be based on finite-dimensional problems.

\section{$p$-SCLI-$n$ for $n$-player Games}
\label{sec:lower-spectral}
\subsection{Definitions and examples}
Let $\cQ^{d_1,...,d_n}$ denote the set of $n$-player quadratic games, i.e. games comprised of $n$ quadratic objectives $l_i : \reals^{d_i} \rightarrow \reals$, and $f_{\mat{A}, \bb} (\bw) \in \cQ^{d_1,...,d_n}$ be a game with vector field $\mat{A}\bw + \bb$ as indicated in eq. \ref{eq:quad-vec-field-n}. The following definition is a direct generalisation of the definition of $p$-SCLI algorithms given by \citet{arjevani2016lower} to $n$-player games.

\begin{definition} [$p$-SCLI-$n$ optimisation algorithms for $n$-player games] \label{def:p-SCLI-n}
Let $\cA$ be an optimisation algorithm for $n$-player quadratic games. Then $\cA$ is a $p$-stationary canonical linear iterative method for $n$-player games ($p$-SCLI-$n$) if there exist functions $C_0, ..., C_{p-1}, N$ from $\reals^{d\times d}$ to $\reals^{d\times d}$-valued random variables, such that the following conditions are satisfied for all $f_{\mat{A},\bb}(\bw)\in\cQ^{d_1,...,d_n}$:
\begin{enumerate}
    \item Given an initialisation $\bw^0,\dots,\bw^{p-1}\in\reals^{d}$, the update rule at iteration $t \geq p$ is given by \begin{align}
    \bw^t = \sum_{i=0}^{p-1} C_{i}(\mat{A}) \bw^{t-p+i}+ N(\mat{A})\bb \label{eq:p-SCLI-n_update_rule}
\end{align}
    \item $C_0(\mat{A}), ..., C_{p-1}(\mat{A}), N(\mat{A})$ are independent from previous iterations
    \item $\EV C_i(\mat{A})$ are finite and simultaneously triangularisable
\end{enumerate}
We will refer to the $C_i$ as the coefficient matrices and $N$ as the inversion matrix.
\end{definition}

An important fact is that if $n = 1$, this definition becomes the same as the one given by \citet{arjevani2016lower}. A key difference, however, is that the Jacobian $\mat{A}$ will generally not be symmetric for $n > 1$; only the blocks $\mat{M}_{ii}$ will be, and hence we may not assume the spectrum $\sigma(\mat{A})$ to be positive since it will generally be complex. Fortunately, several results from \citet{arjevani2016lower} hold nevertheless, as discussed in Appendix \ref{apdx:proof_spectral_lower_bound}. Before introducing the results, let us give examples of algorithms used to optimise games that are $p$-SCLI-$n$, as evidenced by their update rule on quadratic games.
\par
\textbf{Simultaneous Gradient Descent (GD)} The update rule is given by $\bw_i^t = \bw_i^{t-1} - \eta_i \nabla_{\bw_i} l_i(\bw^{t-1})$, which can be rewritten with $\mat{\eta} = \text{Diag}\lrpar{\eta_1,...,\eta_n}$ as: 
\begin{align}
    \bw^t &= \bw^{t-1} - \mat{\eta} \lrpar{\mat{A}\bw^{t-1} + \bb} \nonumber \\
    &= (\mat{I} - \mat{\eta A}) \bw^{t-1} - \mat{\eta}\bb
\end{align}
This shows that simultaneous gradient descent is a $1$-SCLI-$n$ algorithm. 
\par
\textbf{Simultaneous Momentum GD} The update rule is $\bw_i^t = \bw_i^{t-1} - \eta_i \nabla_{\bw_i} l_i(\bw^{t-1}) + \beta_i (\bw_i^{t-1} - \bw_i^{t-2})$ which can rewritten with $\mat{\beta} = \text{Diag}\lrpar{\beta_1,...,\beta_n}$ and $\mat{\eta}$ as before:
\begin{align}
    \bw^t &= \bw^{t-1} - \mat{\eta} \lrpar{\mat{A}\bw^{t-1} + \bb} + \mat{\beta} (\bw^{t-1} - \bw^{t-2}) \nonumber \\
    &= (\mat{I} - \mat{\eta A} + \mat{\beta}) \bw^{t-1} - \mat{\beta} \bw^{t-2} - \mat{\eta}\bb
\end{align}
Therefore, simultaneous gradient descent with momentum is a $2$-SCLI-$n$, if we assume $\mat{\beta}$ to be scalar (since we need the coefficient matrices $C_i(\mat{A})$ to be simultaneously triangularisable).
\par
\textbf{Extragradient }\citep{korpelevich1976extragradient} The update rule is $\bw_i^t = \bw_i^{t-1} - \eta_i \nabla_{\bw_i} l_i(\bw^{t-1} - \mat{\eta}\vec{v}(\bw^{t-1}))$, which can be rewritten as:
\begin{align}\label{eq:EG_update_rule}
    \bw^t &= \bw^{t-1} - \mat{\eta}(\mat{A}(\bw^{t-1} - \mat{\eta}(\mat{A}\bw^{t-1} + \bb)) + \bb) \nonumber \\
    &= (\mat{I -\mat{\eta A} + (\eta A)}^2) \bw^{t-1} - (\mat{I - \eta A})\mat{\eta} \bb
    %\bw^t &= (\mat{I + \eta A})^{-1} (\mat{I + (\eta A)^2}) \bw^{t-1} - (\mat{I + \eta A})^{-1}(\mat{I - \eta A})\mat{\eta} \bb
\end{align}
This shows that extragradient is a $1$-SCLI-$n$. %if $(\mat{I + \eta A})^{-1}$ exists.
\par
\textbf{Simultaneous Stochastic Gradient Descent} The reasoning is the same as the one presented by \citet{arjevani2016lower}: we approximate $\nabla f_{\mat{A}, \bb} (\bw) = \mat{A}\bw + \bb$ with stochastic gradients $\vec{G}_\omega (\bw)$ and denote the error by $\vec{e}_\omega (\bw) = \vec{G}_\omega(\bw) - (\mat{A}\bw + \bb)$. Then the update rule for fixed $\mat{\eta}$ is given by
\begin{align}
    \bw^t &= \bw^{t-1} - \mat{\eta} \vec{G}_{\omega_{t-1}}(\bw^{t-1}) \nonumber \\
    &= (\mat{I} - \mat{\eta A}) \bw^{t-1} - \mat{\eta}\bb - \mat{\eta}\vec{e}_{\omega_{t-1}}(\bw^{t-1})
\end{align}
Under certain assumptions, e.g. if $\vec{e}_{\omega}(\bw) = \mat{A}_\omega \bw + \mat{N}_\omega \bb$ and $\EV\mat{A}_\omega = \EV\mat{N}_\omega = 0$, then the update rule becomes
\begin{align}
    \bw^t = (\mat{I} - \mat{\eta} (\mat{A} + \mat{A}_{\omega_{t-1}})) \bw^{t-1} - \mat{\eta} (\mat{I} + \mat{N}_{\omega_{t-1}})\bb
\end{align}
and we get a $1$-SCLI-$n$.

One last definition is required before we introduce the $p$-SCLI lower bounds. Our definition generalises that of \citet{arjevani2016lower}.

\begin{definition} [Consistency of $p$-SCLI-$n$ optimisation algorithms] \label{def:consistency}
Let $\cQ_{\mat{A}}^{d_1,...,d_n} \subseteq \cQ^{d_1,...,d_n}$ denote the set of quadratic $n$-player games with non-singular Jacobian $\mat{A}$ (see eq. \ref{eq:quad-vec-field-n}). Then $\cA$ is consistent with respect to $\mat{A}$ if for any game $f_{\mat{A},\bb} \in \cQ_{\mat{A}}^{d_1,...,d_n}$ and any initialisation, $\cA$ converges to a stationary point of $f_{\mat{A},\bb}$ or equivalently if the sequence of iterates $(\bw^t)$ (see eq. \ref{eq:p-SCLI-n_update_rule}) satisfies
\begin{align}\label{eq:conv-consistency}
	\bw^t\to-\mat{A}^{-1}\bb
\end{align} 
\end{definition}

Equivalently, as \citet{arjevani2016lower} argue in their section 3.1, consistency with respect to some invertible Jacobian $\mat{A}$ is equivalent to having $\cA$ converge on $f_{\mat{A},\bb}$ and
\begin{align} \label{eq:consistency_sum} 
\sum_{i=0}^{p-1} \EV C_{i}(A) = I_d  + \EV N(\mat{A})\mat{A}
\end{align}
Note that all three examples of optimisation algorithms discussed in this subsection satisfy eq. \ref{eq:consistency_sum}.

\subsection{Parametric lower bound for $p$-SCLI-$n$ with scalar inversion matrix}
We are now ready to introduce the lower bound for $p$-SCLI-$n$ methods with scalar inversion matrix.
\begin{prop}\label{prop:spectral_lower_bound}
Let $\cA$ be a $p$-SCLI-$n$ algorithm with scalar inversion matrix for optimising games over $\reals^{d_1} \times ... \times \reals^{d_n}$. Then for quadratics $f_{\mat{A}, \bb}  \in \cQ^{d_1,...,d_n}$, if $\cA$ is consistent with respect to $\mat{A}$ and if $0 \notin \sigma(\mat{A})$, we have the following lower bound on the (linear) rate of convergence $\rho$:
\begin{align} \label{eq:p-SCLI-n_lower_bound}
    \rho \ge \frac{\sqrt[p]{\kappa}-1}{\sqrt[p]{\kappa}+1} = 1 - \frac{2}{\sqrt[p]{\kappa}+1}
\end{align}
where the condition number $\kappa$ is defined as $\kappa \triangleq \frac{\max |\sigma(\mat{A})|}{\min |\sigma(\mat{A})|}$ where $\sigma(\mat{A})$ is the spectrum of $\mat{A}$.
\end{prop}

While this lower bound is valid for $p$-SCLI-$n$ methods on \emph{any single quadratic game where the assumptions apply}, it also gives us a more general result for problem classes containing quadratics as an immediate corollary, similarly to the lower bounds of the previous section.

\begin{thm}\label{thm:p-SCLI-n-bound}
For any problem class containing quadratic games, there exists a game such that the lower bound given in eq. \ref{eq:p-SCLI-n_lower_bound} holds for a $p$-SCLI-$n$ method.
\end{thm}

\begin{cor}[Iteration complexity bound]
For the same problem classes and under the same assumptions as Theorem \ref{thm:p-SCLI-n-bound}, the minimal number of steps $t$ to satisfy $\max_{i=0,\dots,p-1} \norm{\EV\bw^{t+i} - \EV\bw^*} < \epsilon$ is given by
\begin{align}
    t \geq \lrpar{\frac{\sqrt[p]{\kappa}-1}{2}} \log\lrpar{\frac{C}{\epsilon}}
\end{align}
for some strictly positive constant $C$.
\end{cor}

See Appendix \ref{apdx:proof_spectral_lower_bound} for the proof of Prop. \ref{prop:spectral_lower_bound}, and Appendix \ref{apdx:iteration-complexity} for the iteration complexity bound's proof. Interestingly, by setting $n = 1$, this bound captures the 1-player case for $\mu$-strongly convex objectives with $L$-Lipschitz gradients, where $\kappa = \frac{\max \sigma(\mat{A})}{\min \sigma(\mat{A})} = L/\mu$, verifying the intuition discussed at the end of the previous section, and showing that this technique yields a tight lower bound for $n=1$. Moreover, this form is valid for $n$-player games (and min-max problems) in finite dimension, and $\kappa$ arises naturally from the spectral properties of the update rules of the $p$-SCLI-$n$ methods and is lower bounded by 1. Additionally, our bounds are valid for some stochastic methods. In single-objective optimisation (i.e. $n=1$), while linear rates are not achievable for general stochastic problems, for which the worst-case bounds are sublinear, under certain conditions linear rates are possible \citep{loizou2017momentum}. Conditions of this type can be satisfied in over-parameterised neural networks \citep{vaswani2019convergence}. Hence, our linear lower bounds may be useful even for stochastic problems.

However, while the moduli in the $n>1$ case allow us to handle complex spectra and matches the classical definition of condition number from linear algebra, several analyses have shown that not only the modulus, but also the relative size of the real and imaginary parts of elements of the spectrum matter \citep{mescheder2017numerics,gidel2019negative}. Such an analysis may yield more expressive bounds, but is out of the scope of this work. We will nevertheless give a more explicit form of the bound for 2-player games for which $d_1=d_2$ that will make the $\mu_i$ and $L_i$ appear.

% Note that in Proposition~\ref{prop:spectral_lower_bound}, we provide a lower bound for the convergence rate to limit points of the dynamics which may not be Nash equilibria. Assuming the existence of a Nash equilibrium would lead to a smaller class of matrices $\mat{A}$ (where $\mat{M}_{ii} \succeq 0$ in~eq.~\ref{eq:quad-vec-field-n}) and potentially to a tighter lower bound.

\textbf{Some explicit bounds for $p$-SCLI-2 with $d_1 = d_2$} Prop. \ref{prop:spectral_lower_bound} may be used to derive lower bounds for $2$-player games for which $d_1 = d_2$. These bounds depend on the value of the $\mu_i$ and $L_i$ defined as in eq. \ref{eq:def_spectrum_bounds}. Namely, let
 \begin{alignat}{2}
&\Delta_\mu &&= (\mu_1 + \mu_2)^2 - 4(\mu_1\mu_2 + \mu_{12}^2) \nonumber \\
& &&= (\mu_1 - \mu_2)^2 - 4 \mu_{12}^2\\
&\Delta_L  &&= (L_1 + L_2)^2 - 4(L_1L_2 + L_{12}^2) \nonumber \\
& &&= (L_1 - L_2)^2 - 4 L_{12}^2
 \end{alignat}
Table \ref{table:p-SCLI-2_kappa_lower_bounds} gives lower bounds on the condition number that may then be plugged into eq. \ref{eq:p-SCLI-n_lower_bound} to get lower bounds on two-players games corresponding to min-max problems (and are therefore lower bounds for general 2-player games).
\begin{table}[H]
\centering
\caption{Lower Bounds on the Condition Number}
\begin{tabularx}{0.43\textwidth}{@{}lcc@{}} 
\toprule
                      & $\Delta_\mu < 0$                 & $\Delta_\mu \geq 0$                                                         \\
\midrule
$\Delta_L < 0$ & $\kappa = \sqrt{\frac{L_1L_2 + L_{12}^2}{\mu_1\mu_2 + \mu_{12}^2}}$ & $\kappa \geq 2\frac{\sqrt{L_1L_2 + L_{12}^2}}{\mu_1+\mu_2 - \sqrt\Delta_\mu}$                          \\ 
$\Delta_L \geq 0$         & $\kappa \geq \frac{1}{2}\frac{L_1+L_2 + \sqrt\Delta_L}{\sqrt{\mu_1\mu_2 + \mu_{12}^2}}$& $\kappa \geq \frac{L_1+L_2 + \sqrt\Delta_L}{\mu_1+\mu_2 - \sqrt\Delta_\mu}$ \\
\bottomrule
\end{tabularx}\label{table:p-SCLI-2_kappa_lower_bounds}
\end{table}

See Appendix \ref{apdx:p-SCLI-2-refined-lower} for the counterexample in \ref{eq:minmax} leading to these bounds. This result resolves the issues raised in the discussion of the domino bounds as it uses all of the $\mu_i$ and $L_i$, and is proven from finite-dimensional problems. In fact, if one sets $\mu_1 = L_1$ and $\mu_2 = L_2$ such that $\mu_1$ and $\mu_2$ are small (in particular, smaller than $\mu_{12}$) and $L_{12}$ is large (in particular, larger than $L_1, L_2$), table \ref{table:p-SCLI-2_kappa_lower_bounds} yields $\kappa = \sqrt{\frac{\mu_1\mu_2 + L_{12}^2}{\mu_1\mu_2 + \mu_{12}^2}}$, which coincides with the $\kappa$ from Thm. \ref{thm:domino_good_lower_bound_infdim}. More generally, for $p = 1$, if both $\Delta_L$ and $\Delta_{\mu}$ are negative, we get a tighter bound from the $p$-SCLI-2 formalism for $1$-SCLI-2 methods that also satisfy the two-step linear span assumption than from Thm. \ref{thm:domino_good_lower_bound_infdim}. Finally, it provides the same plug-and-play convenience as $p$-SCLI to derive bounds for a large class of algorithms that may not satisfy the first-order black box assumption. On the other hand, the bounds may not be tight for $p \geq 3$, as it was the case for single-objective $p$-SCLI \citep{arjevani2016lower}.

An interesting case is $p = 2$: for $2$-SCLI-$2$ methods that satisfy the two-step linear span assumption such as negative momentum, the rate stemming from the $p$-SCLI-2 analysis appears to be smaller than the rate from the improved domino bound. This may be because the proof techniques used in our generalisation of $p$-SCLI yield bounds that can be improved for $p>1$. In particular, a key difference between Thm. \ref{thm:domino_good_lower_bound_infdim} and \ref{thm:p-SCLI-n-bound} is that as explained in the background section, lower bounds generally need not apply to every single problem within a class. However, due to the proof technique used, the bound given in Prop. \ref{prop:spectral_lower_bound} has to apply to every single quadratic game where the assumptions hold. Because some games might be inherently easier than others, such games may bottleneck the bound in Prop. \ref{prop:spectral_lower_bound} and Thm. \ref{thm:p-SCLI-n-bound}, and introduce looseness. Indeed, first, the number of players does not appear in Prop. \ref{prop:spectral_lower_bound}'s proof, meaning that the proof yielded a bound that should hold for $n=1$ which might inherently have faster convergence than $n>1$ due to having to optimise only one objective over one set of parameters. Second, in the proof, because we do not impose a structure for the quadratic games, the bound has to hold for purely cooperative games (that is, the objective of each player does not depend on other players' parameters), in which case the situation is analogous to $n$ single-objective optimisation problems, where we may expect the convergence to be faster than in games with interactions between players. As such, it is worth considering whether fixing the number of players and the structure of the games earlier in the proof could improve the bound.

\section{Conclusion}
In this work, we provide linear lower bounds and condition numbers for any problem class containing quadratic or bilinear games. We give a lower bound on the rate of convergence of first-order black box methods for $2$-player games (which directly applies to min-max and saddle point problems) satisfying the two-step linear span assumption by generalising Nesterov's lower bound for the optimisation of strongly convex, smooth convex objectives to 2-player games (R.Q. \ref{rq:1}) and constructing a novel class of hard problems using spectral properties of a class of operators in Banach algebras. Moreover, we generalise the framework of $p$-SCLI, which requires symmetricity of the Hessian in single-objective optimisation, to provide a bound for a large class of optimisers for $n$-player games by extending the results of $p$-SCLI to quadratic games with non-symmetric Jacobian, which for $n=1$ recovers \cite{arjevani2016lower}'s tight bounds. We then give explicit bounds for 2-player games, which apply to min-max and saddle point problems (R.Q. \ref{rq:2}). Finally, we derived formulations for the condition number that matched (in the case of the first domino bound), or were more general (in the case of the improved domino bound, and $p$-SCLI-$n$ and $p$-SCLI-2 bounds) than the existing ones in the upper bound literature (R.Q. \ref{rq:3}). As in the single-objective case, our bounds and condition numbers suggest that optimisers may converge faster on games for which the eigenvalues are at a similar, remote distance from the origin (e.g. on a circle) than on games for which some eigenvalues are close to and others are far from 0.

Following the initial release of this work, several other authors have built upon the bounds presented in this paper. The bound from Theorem \ref{thm:domino_good_lower_bound_infdim} is tight on the class of smooth strongly-convex-strongly-concave games, as it is matched by the upper bound presented by \citet{fallah2020optimal}, and up to logarithmic factors, by the upper bound of \citet{lin2020near}. Additionally, this same lower bound was also shown to be tight on the class of bilinear games with non-singular Jacobian, which are merely convex-concave, by Prop. 4 of \citet{azizian2020accelerating}. As a corollary, our results establish the optimality of those methods on the aforementioned classes of problems.

However, several directions remain to be explored. For example, we raised the question of whether the $p$-SCLI-$n$ bound could be tightened for $p>1$ by improving the proof technique, especially given that we are not aware of faster rates of convergence in the literature than those of $p=1$ or those of the improved domino bound. Moreover, we would like to present a more exhaustive overview of the extension of $p$-SCLI, and discuss the resulting upper and lower bounds for various commonly used algorithms. In particular, we would like to extend our lower bounds to $p$-SCLI-$n$ with diagonal inversion matrices, as \citet{arjevani2016lower} did in the $p$-SCLI framework, and provide bounds in the 2-player case when $d_1 \neq d_2$. Furthermore, we believe tighter bounds may be derived, for example by adding constraints on the $C_i$ or by looking not only at the modulus of the eigenvalues but also at their arguments, as done by \citet{gidel2019negative}, since we know that the relative size of the imaginary part and real part (even at fixed modulus) affects the dynamics in games \citep{mescheder2017numerics}. Finally, it would be important to understand when and how linear convergence is possible in non strongly-convex-strongly-concave settings. We plan on exploring several of these directions in future work. 

\section*{Acknowledgements}

The authors would like to thank Damien Scieur for useful discussions and feedback.
This work was partially supported by the FRQNT new researcher program (2019-NC-257943), the NSERC Discovery grant (RGPIN-2019-06512), a startup grant by IVADO, a Microsoft Research collaborative grant and a Canada CIFAR AI chair.
Gauthier Gidel was partially supported by a Borealis AI Graduate Fellowship.

% This work was supported by the FRQNT new researcher program (2019-NC-257943), the NSERC Discovery grant (RGPIN-2019-06512), a startup grant by IVADO and a Canada CIFAR AI chair. 

\bibliographystyle{plainnat}
\bibliography{biblio}

\onecolumn
\begin{appendices}
\section{$n$-player quadratic games}\label{apdx:n-player-games}
While our work focuses mostly on $2$-player games, one of the main results, Prop. \ref{prop:spectral_lower_bound}, is independent of the number of players, and is proven for general $n$. Therefore, a short discussion of the form of quadratic $n$-player games is provided below. 

For an $n$-player game, the general form of a quadratic is given by
\begin{align}
    l_i(\bw) = \sum_{j=1}^n \sum_{k=1}^n \bw_j^\top \mat{M}_{ijk} \bw_k + \sum_{j=1}^n \bw_j^\top \bb_{ij} + c_i \,.
\end{align}
Because the dynamics depend only on the $\nabla_{\bw_i} l_i(\bw)$, we will get equivalent dynamics by pruning the terms that do not depend on $\bw_i$ and working directly with the simpler objectives
\begin{align}
    l_i(\bw) &= \frac{1}{2} \bw_i^\top \mat{M}_{ii} \bw_i + \sum_{j\neq i}^n \bw_i^\top \mat{M}_{iij} \bw_j + \sum_{j\neq i}^n \bw_j^\top \mat{M}_{iji} \bw_i + \bw_i^\top \bb_{ii} \nonumber \\
            &= \frac{1}{2} \bw_i^\top \mat{M}_{ii} \bw_i + \sum_{j\neq i}^n \bw_i^\top \mat{M}_{ij} \bw_j + \bw_i^\top \bb_{i} 
\end{align}
where we have let $\mat{M}_{ij} \triangleq \mat{M}_{iij} + \mat{M}_{iji}^\top, \bb_i \triangleq \bb_{ii}\,,\, 1\leq i,j\leq n$. Note that we may assume the $\mat{M}_{ii}$ to be symmetric, since in general $\bx^\top \mat{A} \bx = \frac{1}{2} \bx^\top (\mat{A} + \mat{A}^\top) \bx$. Thus, we can write:
\begin{align}
    \nabla_{\bw_i}l_i(\bw) = \begin{pmatrix}
    \mat{M}_{i1} & \dots & \mat{M}_{in}
    \end{pmatrix}
    \bw + \bb_i
\end{align}
which yields the following equation for the vector field:
\begin{align}
    \bv(\bw) = \mat{A}\bw + \bb, \quad\quad \mat{A} \triangleq \begin{pmatrix}
 \mat{S}_{1} & \dots & \mat{M}_{1n}\\
 \vdots & & \vdots \\ 
 \mat{M}_{n1} &\dots & \mat{S}_{n}
\end{pmatrix}, 
\quad \bb \triangleq \begin{pmatrix}
\bb_1 \\
\vdots \\
\bb_n
\end{pmatrix}
\end{align}
%\begin{align} \label{eq:jacobian-n-quadratic}
%    \mat{J}(\bv) = \begin{pmatrix}
% \mat{M}_{11} & \dots & \mat{M}_{1n}^\top\\
% \vdots & & \vdots \\ 
% \mat{M}_{n1}^\top &\dots & \mat{M}_{nn}
%\end{pmatrix}
%\end{align}
%where we have used $\mat{M}_{ii} = \mat{M}_{ii}^\top$.

where $\mat{A}$ is the Jacobian of $\bv$ and where we let $\mat{S}_i \triangleq \mat{M}_{ii}$.

\newpage
\section{Proofs of Nesterov's bounds for games}\label{apdx:domino_lower_bound_infdim_proof}
The proofs in this section are based on min-max problems for a class of functions\footnote{As evoked in the discussion of the improved bound, for these functions, a limitation is that $\mu_1 = L_1$ and $\mu_2 = L_2$, but we were not able to find counterexamples with $L_i \neq \mu_i$ for which the bound had simple enough closed form, even when choosing terms of the form $\bx^\top \mat{M_{\bx}} \bx, \by^\top \mat{M_{\by}} \by$ with $\mat{M_{\bx}}, \mat{M_{\by}}$ bidiagonal or tridiagonal.} $f:\ell_2 \times \ell_2 \rightarrow \mathbb{R}$ such that
\begin{align}
    f(\bx, \by) = c\bx^\top \mat{M}\by - d_1 \bx^\top \vec{e_1} + d_2 \by^\top \vec{e_1} + \frac{\mu_1}{2} \norm{\bx}^2 - \frac{\mu_2}{2} \norm{\by}^2
\end{align}
where $\vec{e_1}$ is a vector with a $1$ in the first entry and $0$ elsewhere, $c, d_1, d_2 \in \mathbb{R}$, and $\mu_1, \mu_2 \in \mathbb{R}^{++}$, with $\mat{M}$ upper bidiagonal matrix such that
\begin{align}
    \mat{M} = \begin{bmatrix}
        a_0 & a_1 & 0 & 0 & \dots\\
        0 & a_0 & a_1 & 0 & \dots\\
        0 & 0 & a_0 & a_1 & \dots\\
        \vdots & & & \ddots & \ddots
    \end{bmatrix}
\end{align}
where $a_0, a_1 \neq 0$. 

As \citet{nesterov2013introductory}, we shall assume that $\bx_0, \by_0$ are initialised at 0, as otherwise we may work with $\bx - \bx_0$ and $\by - \by_0$ in the counterexample and perform the change of variable $\bx \leftarrow \bx - \bx_0, \, \by \leftarrow \by - \by_0$ (which would give us zero-initialisation)  and switch back at the end of the analysis.

\subsection{On the domino argument}\label{apdx:domino_argument}
More about the domino argument can be found in \citet{nesterov2013introductory}; here, we shall give the intuition as to why it works. Let us introduce the ingredients of the update rule under our assumptions.
\begin{align}
    \mat{A} = \begin{pmatrix}
    \mu_1 & \mat{M} \\
    -\mat{M}^\top & \mu_2
    \end{pmatrix} \quad\quad
    \mat{A}^2 = \begin{pmatrix}
    \mu_1^2 - \mat{MM}^\top & (\mu_1 + \mu_2) \mat{M} \\
    -(\mu_1 + \mu_2) \mat{M}^\top & \mu_2^2 -\mat{M}^\top\mat{M}
    \end{pmatrix} \quad \quad
    \bb = \begin{pmatrix}
    -d_1 \vec{e}_1 \\
    d_2 \vec{e}_2 \end{pmatrix} \\
    \mat{A}\bb = \begin{pmatrix}
    -\mu_1 d_1 \vec{e}_1 + d_2 \mat{M}\vec{e}_1 \\
    d_1 \mat{M}^\top \vec{e}_1 + \mu_2 d_2 \vec{e}_1
    \end{pmatrix} \quad\quad
    \mat{M}\vec{e}_1 = \begin{pmatrix} a_0 & 0 & \dots \end{pmatrix}^\top \quad\quad
    \mat{M}^\top\vec{e}_1 = \begin{pmatrix} a_0 & a_1 & 0 & \dots \end{pmatrix}^\top \\
    \mat{A}\bw = \begin{pmatrix}
    \mu_1 \bx + \mat{M} \by \\
    -\mat{M}^\top \bx + \mu_2 \by 
    \end{pmatrix} \quad\quad
    \mat{A}^2 \bw = \begin{pmatrix}
    (\mu_1^2 - \mat{MM}^\top)\bx + (\mu_1+\mu_2)\mat{M}\by \\
    -(\mu_1+\mu_2)\mat{M}^\top \bx + (\mu_2^2 - \mat{M}^\top\mat{M})\by 
    \end{pmatrix}
\end{align}

\subsubsection{One-step linear span assumption}
If the algorithm follows the one-step assumption (e.g. gradient descent), which we define as 
\begin{align}
    \bw_t \in \bw_0 + \text{Span} \big(\bw_0, ..., \bw_{t-1}, \mat{A}\bw_0, ..., \mat{A}\bw_{t-1}, \bb \big)
\end{align}
note that the part of $\bb$ contributing to the update rules of both $\bx$ and $\by$ will have be a vector with a single non-zero entry as its first entry, i.e. $(*, 0, ...)$. Therefore,
\begin{align}
    \bx_t \in \text{Span}\big((*, 0, ...), \bx_0, ..., \bx_{t-1}, \mat{M} \by_0, ..., \mat{M} \by_{t-1}\big) \\
    \by_t \in \text{Span}\big((*, 0, ...), \by_0, ..., \by_{t-1}, \mat{M}^\top \bx_0, ..., \mat{M}^\top \bx_{t-1} \big)
\end{align}
Since $(\bx_0, \by_0) = 0$ but $(\bx^*, \by^*) \neq 0$, we want to see, at every iteration $t$, how many components of $\bx_t, \by_t$ have been \emph{initialised}, i.e. received information from (which components of) past iterations and therefore could have changed from their initial values of zero. The dependence of the components of $\bx_t$ and $\by_t$ on past iterates, based on the one-step linear span assumption, is summarised below:
\begin{align}
    \text{Comp. $i = 1$ of $\bx_t$}&\text{: comp.~1 from const. vector, $i$ from $\bx_0,...,\bx_{t-1}$, and $i, i+1$ from $\by_0,...,\by_{t-1}$} \nonumber \\
    \text{Comp. $i = 1$ of $\by_t$}&\text{: comp.~1 from const. vector, $i$ from $\by_0,...,\by_{t-1}$, and $i$ from $\bx_0,...,\bx_{t-1}$} \nonumber \\
    \text{Comp. $i\geq2$ of $\bx_t$}&\text{: comp.~$i$ from $\bx_0,...,\bx_{t-1}$, and $i, i+1$ from $\by_0,...,\by_{t-1}$} \nonumber \\
    \text{Comp. $i\geq2$ of $\by_t$}&\text{: comp.~$i$ from $\by_0,...,\by_{t-1}$, and $i-1, i$ from $\bx_0,...,\bx_{t-1}$} \nonumber 
\end{align}

Therefore, if $(\bx_0, \by_0) = 0$, it is clear that the only terms in the update rule that may initialise any new component of $(\bx_1, \by_1)$ are the constant vectors. Thus, in $(\bx_1, \by_1)$ only the first component will be initialised if we are using simultaneous first-order black box methods satisfying the one-step linear span assumption, and additionally the second component of $\by_1$ if we extended the definition of the linear span assumption to use $\bx_t$ instead of $\bx_{t-1}$ when computing $\by_t$. 

We then move on to $(\bx_2, \by_2)$ and compute from the rules above which components, i.e. values of $i$, can be initialised given the initialisation of the past iterates. For simultaneous methods, we see that we still cannot initialise the second component of $\bx_2$ since that would require the second component of $\bx_0$ or $\bx_1$, or the second or third components of either $\by_0$ or $\by_1$ to have been initialised. Nevertheless, given that the second component of $\by_2$ depends on the first component of $\bx_0, \bx_1$, we may initialise a second component in $\by_2$. However, a third component would require either the third component of $\by_0, \by_1$ or the second or third components of $\bx_0, \bx_1$ to be already initialised, which is not the case. Therefore, in simultaneous one-step methods, only 1 component of $\bx_2$ and 2 components of $\by_2$ will be initialised at most.

This logic is applied in table \ref{table:domino_argument}, which indicates the number of components in both sets of parameters that have been updated from their initial value (e.g. that are nonzero if we initialise the parameters at 0) at each iteration.

\begin{table}[H] 
		\caption{Number of components initialised in $\bx_t$ and $\by_t$ at iteration $t$, for methods using $\bw_i, \mat{A}\bw_i, \bb$} \label{table:domino_argument}
		\centering
    \begin{tabularx}{0.83\textwidth}{@{}lcccccc@{}} 
    \toprule
        Iteration &\multicolumn{2}{c}{Simultaneous} & \multicolumn{2}{c}{Alt. $\bx_t$ instead of $\bx_{t-1}$ for $\by_t$} & 
        \multicolumn{2}{c}{Alt. $\by_t$ instead of $\by_{t-1}$ for $\bx_t$}\\
        \cmidrule(lr){2-3} \cmidrule(lr){4-5} \cmidrule(lr){6-7} t & \# dim $\bx_t$ & \# dim $\by_t$ & \# dim $\bx_t$ & \# dim $\by_t$ & \# dim $\bx_t$ & \# dim $\by_t$ \\
		\midrule
		0 & 0 & 0 & 0 & 0 & 0 & 0\\
		1 & 1 & 1 & 1 & 2 & 1 & 1\\
		2 & 1 & 2 & 2 & 3 & 2 & 2\\
		3 & 2 & 2 & 3 & 4 & 3 & 3\\
		4 & 2 & 3 & 4 & 5 & 4 & 4\\
		\bottomrule
		\end{tabularx}
\end{table}

A simple proof by induction can generalise that for both alternating or simultaneous updates, at most $t+1$ components of $\bx_t, \by_t$ have been initialised. The consequence is that at iteration $t$ we have $\bx_t(i) = \bx_0(i), \by_t(i) = \by_0(i)$ for $i > t+1$, where $(\bx_0, \by_0) = 0$. Note that this still holds if we compute elements of the span with diagonal matrices as coefficients. This can be summarised as the following.
\begin{lemma}[One-step linear span domino argument]
Suppose $(\bx_0, \by_0) = 0$. Then for algorithms satisfying the one-step linear span assumption (where elements of the span may be computed using diagonal matrices as coefficients), we have
\begin{align}
    \begin{matrix}
    \bx_t(i) = 0\\
    \by_t(i) = 0
    \end{matrix} \quad \quad \textit{for } i > t+1
\end{align}
\end{lemma}

\subsubsection{Two-step linear span assumption}
For an algorithm satisfying the two-step assumption such as extragradient (see eq. \ref{eq:EG_update_rule} for the update rule), i.e. if we have
\begin{align}
    \bw_t \in \bw_0 + \text{Span} \big(\bw_0, ..., \bw_{t-1}, \mat{A}\bw_0, ..., \mat{A}\bw_{t-1}, \mat{A}^2 \bw_0, ..., \mat{A}^2 \bw_{t-1}, \bb, \mat{A}\bb\big)
\end{align}
the part of $\bb$ and $\mat{A}\bb$ contributing to the update rule on $\bx$ will be a vector of the form $(*, 0, ...)$ and the part contributing to the update rule on $\by$ will have the form $(*, *, 0, ...)$. Therefore, 
\begin{align}
    \bx_t \in \text{Span}\big((*, 0, 0, ...), \bx_0, ..., \bx_{t-1}, \mat{M} \by_0, ..., \mat{M} \by_{t-1}, \mat{MM}^\top \bx_0, ..., \mat{MM}^\top \bx_{t-1} \big) \\
    \by_t \in \text{Span}\big((*, *, 0, ...), \by_0, ..., \by_{t-1}, \mat{M}^\top \bx_0, ..., \mat{M}^\top \bx_{t-1}, \mat{M}^\top\mat{M} \by_0, ..., \mat{M}^\top\mat{M} \by_{t-1} \big)
\end{align}
We can see from eq. \ref{eq:MM_trans} ($\mat{M}^\top\mat{M}$ yields the same matrix with $a_0^2$ instead of $a_0^2 + a_1^2$ in the first entry) that the $i$-th component of $\mat{MM}^\top \bx$ depends on the $i-1, i, i+1$-th components of $\bx$ for $i \geq 2$. Since only the number of initialised components will interest us, we want to see, at every iteration $t$, how many components of $\bx_t, \by_t$ received information from past iterations and therefore could have changed from their initial values of zero. The dependence of the components of $\bx_t$ and $\by_t$ on past iterates, based on the two-step linear span assumption, is summarised below:
\begin{align}
    \text{Comp. $i = 1$ of $\bx_t$}&\text{: comp.~1 from const., $i, i+1$ from $\bx_0,...,\bx_{t-1}$, and $i, i+1$ from $\by_0,...,\by_{t-1}$} \nonumber \\
    \text{Comp. $i = 1$ of $\by_t$}&\text{: comp.~1 from const., $i, i+1$ from $\by_0,...,\by_{t-1}$, and $i$ from $\bx_0,...,\bx_{t-1}$} \nonumber \\
    \text{Comp. $i = 2$ of $\bx_t$}&\text{: comp.~$i-1,i,i+1$ from $\bx_0,...,\bx_{t-1}$, and $i, i+1$ from $\by_0,...,\by_{t-1}$} \nonumber \\
    \text{Comp. $i = 2$ of $\by_t$}&\text{: comp.~1 from const., $i-1,i,i+1$ from $\by_0,...,\by_{t-1}$, and $i-1, i$ }\nonumber \\
    &\text{\ \ from $\bx_0,...,\bx_{t-1}$} \nonumber \\
    \text{Comp. $i > 2$ of $\bx_t$}&\text{: comp.~$i-1,i,i+1$ from $\bx_0,...,\bx_{t-1}$, and $i, i+1$ from $\by_0,...,\by_{t-1}$} \nonumber \\
    \text{Comp. $i > 2$ of $\by_t$}&\text{: comp.~$i-1,i,i+1$ from $\by_0,...,\by_{t-1}$, and $i-1, i$ from $\bx_0,...,\bx_{t-1}$} \nonumber 
\end{align}
so for the first few iterations we get Table \ref{table:domino_argument_extragradient}.

\begin{table}[H] 
		\caption{Number of components initialised in $\bx_t$ and $\by_t$ for methods using $\bw_i, \mat{A}\bw_i, \mat{A}^2\bw_i, \bb, \mat{A}\bb$} \label{table:domino_argument_extragradient}
		\centering
    \begin{tabularx}{0.83\textwidth}{@{}lcccccc@{}} 
    \toprule
        Iteration &\multicolumn{2}{c}{Simultaneous} & \multicolumn{2}{c}{Alt. $\bx_t$ instead of $\bx_{t-1}$ for $\by_t$} & 
        \multicolumn{2}{c}{Alt. $\by_t$ instead of $\by_{t-1}$ for $\bx_t$}\\
        \cmidrule(lr){2-3} \cmidrule(lr){4-5} \cmidrule(lr){6-7} t & \# dim $\bx_t$ & \# dim $\by_t$ & \# dim $\bx_t$ & \# dim $\by_t$ & \# dim $\bx_t$ & \# dim $\by_t$ \\
		\midrule
		0 & 0 & 0 & 0 & 0 & 0 & 0\\
		1 & 1 & 2 & 1 & 2 & 2 & 2\\
		2 & 2 & 3 & 2 & 3 & 3 & 3\\
		3 & 3 & 4 & 3 & 4 & 4 & 4\\
		4 & 4 & 5 & 4 & 5 & 5 & 5\\
		\bottomrule
		\end{tabularx}
\end{table}
Hence, we can prove once again by induction that in any case at iteration $t$ we have $\bx_t(i) = \bx_0(i), \by_t(i) = \by_0(i)$ for $i > t+1$ and $(\bx_0, \by_0) = 0$ for methods also accessing $\mat{A}^2 \bw_i, \mat{A}\bb$. Here again, this still holds if we multiply the entries in our span by diagonal matrices. We can once again summarise this as a lemma.
\begin{lemma}[Two-step linear span domino argument]
Suppose $(\bx_0, \by_0) = 0$. Then for algorithms satisfying the two-step linear span assumption (where elements of the span may be computed using diagonal matrices as coefficients), we have
\begin{align}
    \begin{matrix}
    \bx_t(i) = 0\\
    \by_t(i) = 0
    \end{matrix} \quad \quad \textit{for } i > t+1
\end{align}
\end{lemma}

\subsection{Proof of Prop.~\ref{prop:domino_bad_lower_bound_infdim}}\label{apdx:domino_bad_lower_bound_infdim_proof}
We look for stationary points $(\bx^*, \by^*)$:
\begin{align}
    \nabla_{\bx} f(\bx^*, \by^*) &= c\mat{M}\by^* - d_1 \vec{e_1} + \mu_1 \bx^* = 0\\
    \nabla_{\by} f(\bx^*, \by^*) &= c\mat{M}^\top \bx^* + d_2 \vec{e_1} - \mu_2 \by^* = 0
\end{align}
Therefore, denoting $x_i = \bx^*(i), y_i = \by^*(i)$, the components of stationary points satisfy the recurrence:
\begin{alignat}{2}
    x_1&: a_0c y_1 + a_1 c y_2 &- d_1 + \mu_1 x_1 = 0 \label{eq:ss2_x_1} \\ 
    y_1&: a_0c x_1 &+ d_2 - \mu_2 y_1 = 0 \label{eq:ss2_y_1}
\end{alignat}
and for $n\geq 2$:
\begin{align}
    x_n&: a_0 c y_n + a_1 c y_{n+1} + \mu_1 x_n = 0 \label{eq:ss2_on_x_n} \\
    y_n&: a_1 c x_{n-1} + a_0 c x_n - \mu_2 y_n = 0 \label{eq:ss2_on_y_n}
\end{align}
We can rewrite the above as
\begin{align}
    x_n &= -a_0 \frac{c}{\mu_1} y_n - a_1 \frac{c}{\mu_1} y_{n+1} \label{eq:ss2_x_n}\\
    y_n &= a_1 \frac{c}{\mu_2} x_{n-1} + a_0 \frac{c}{\mu_2} x_n \label{eq:ss2_y_n}
\end{align}
and using eq.~\ref{eq:ss2_y_n} to substitute $y_n$ in eq.~\ref{eq:ss2_on_x_n} we get a recurrence on $x$ only:
\begin{align}
    a_0 a_1 \frac{c^2}{\mu_2} x_{n-1} + a_0^2 \frac{c^2}{\mu_2} x_n + a_1^2 \frac{c^2}{\mu_2} x_n + a_0 a_1 \frac{c^2}{\mu_2} x_{n+1} + \mu_1 x_n = 0
\end{align}
which can be rewritten as
\begin{align}
    a_0 a_1 x_{n+1} + \left( \frac{\mu_1 \mu_2}{c^2} + a_0^2 + a_1^2 \right) x_n + a_0 a_1 x_{n-1} = 0 \label{eq:ss2_clean_rec_x_n}
\end{align}
The roots of the characteristic polynomial of the above linear recurrence are given by
\begin{align}
    \chi_{\pm} = \frac{-\left( \frac{\mu_1 \mu_2}{c^2} + a_0^2 + a_1^2 \right) \pm \sqrt{\left( \frac{\mu_1 \mu_2}{c^2} + a_0^2 + a_1^2 \right)^2 - 4 a_0^2 a_1^2}}{2 a_0 a_1}
\end{align}
and the solution to the linear recurrence is given by $x_n = C_1 \chi_+^n + C_2 \chi_-^n$ (see \cite{brassard1996fundamentals} for a reference on solving linear recurrences). Note that 
\begin{align}
    \chi_{\pm} + 1 &= \frac{-\left( \frac{\mu_1 \mu_2}{c^2} + a_0^2 + a_1^2 \right) \pm \sqrt{\left( \frac{\mu_1 \mu_2}{c^2} + a_0^2 + a_1^2 \right)^2 - 4 a_0^2 a_1^2} + 2 a_0 a_1}{2 a_0 a_1} \nonumber \\
    &= \frac{-\left( \frac{\mu_1 \mu_2}{c^2} + (a_0 - a_1)^2 \right) \pm \sqrt{\left( \frac{\mu_1 \mu_2}{c^2} + a_0^2 + a_1^2 \right)^2 - 4 a_0^2 a_1^2}}{2 a_0 a_1}
\end{align}
Suppose $a_0 a_1 > 0$. As $\frac{\mu_1 \mu_2}{c^2} > 0$, we have $\chi_{-} + 1 < 0$ i.e. $\lvert\chi_{-}\rvert > 1$. Similarly, if we had $a_0 a_1 < 0$ instead, we would have $\chi_{-} - 1 > 0$ which also yields $\lvert\chi_{-}\rvert > 1$. Therefore, $\chi_{-}$ is not a solution as it will not yield a $\bx$ in $\ell_2$. However, note that $\chi_{+} \chi_{-} = 1$ which implies that we always have $\lvert\chi_{+}\rvert < 1$. Hence, we are only concerned with $\chi \triangleq \chi_{+}$. Moreover, note that the square root always exist as we can rewrite the content of the square root to show that it is always positive:
\begin{align}
    \chi &= \frac{-\left( \frac{\mu_1 \mu_2}{c^2} + a_0^2 + a_1^2 \right) + \sqrt{\left( \frac{\mu_1 \mu_2}{c^2} + a_0^2 + a_1^2 \right)^2 - 4 a_0^2 a_1^2}}{2 a_0 a_1} \nonumber \\
    &= \frac{-\left( \frac{\mu_1 \mu_2}{c^2} + a_0^2 + a_1^2 \right)+ \sqrt{\left(\frac{\mu_1 \mu_2}{c^2}\right)^2 + 2 \frac{\mu_1 \mu_2}{c^2}\left(a_0^2 + a_1^2 \right) + \left(a_0^2 + a_1^2 \right)^2 - 4 a_0^2 a_1^2}}{2 a_0 a_1} \nonumber \\
    &= \frac{-\left( \frac{\mu_1 \mu_2}{c^2} + a_0^2 + a_1^2 \right)+ \sqrt{\left(\frac{\mu_1 \mu_2}{c^2}\right)^2 + 2 \frac{\mu_1 \mu_2}{c^2}\left(a_0^2 + a_1^2 \right) + \left(a_0^2 - a_1^2 \right)^2}}{2 a_0 a_1}
\end{align}
In order to simplify the results, we let $a_0 = -a_1 = 1$ and we get:
\begin{align}
    \chi &= \left( \frac{\mu_1 \mu_2}{2c^2} + 1 \right) - \sqrt{\left(\frac{\mu_1 \mu_2}{2c^2}\right)^2 + \frac{\mu_1 \mu_2}{c^2}}
\end{align}
One may note that $L_{12} = c\sqrt{\rho(\mat{MM}^\top)}$. As we have
\begin{align}\label{eq:MM_trans}
    \mat{MM}^\top = \begin{bmatrix}
        a_0^2+a_1^2 & a_0 a_1 & 0 & 0 & \dots\\
        a_0 a_1 & a_0^2+a_1^2 & a_0 a_1 & 0 & \dots\\
        0 & a_0 a_1 & a_0^2+a_1^2 & a_0 a_1 & \dots\\
        \vdots & & \ddots & \ddots & \ddots
    \end{bmatrix}
    = \begin{bmatrix}
        2 & -1 & 0 & 0 & \dots\\
        -1 & 2 & -1 & 0 & \dots\\
        0 & -1 & 2 & -1 & \dots\\
        \vdots & & \ddots & \ddots & \ddots
    \end{bmatrix}
\end{align}
we note that $\mat{MM}^\top$ is a tridiagonal Toeplitz matrix, for which the upper end of the spectrum is given by (see Theorem 7.20 of \citet{douglas2012banach}) 
\begin{align} \label{eq:apdx-domino-spectrum-checkpoint}
    \sup |\sigma(\mat{MM}^\top)| &= \text{ess} \sup_{\theta\in[0, 2\pi)} (a_0 a_1 e^{-i\theta} + (a_0^2+a_1^2) + a_0 a_1 e^{i\theta}) \nonumber \\
    &= \text{ess} \sup_{\theta\in[0, 2\pi)} (2a_0 a_1 \cos\theta + (a_0^2+a_1^2)) = 4
\end{align}
and therefore $L_{12} = 2c$. Defining the condition number as
\begin{align} \label{eq:kappa_bad_bound}
    \kappa = \frac{L_{12}}{\sqrt{\mu_1 \mu_2}}
\end{align}
to retrieve the condition number from the upper bound literature, we get that
\begin{align} \label{eq:xi_bad_bound}
    \chi &= \left( \frac{2}{\kappa^2} + 1 \right) - \sqrt{\frac{4}{\kappa^4} + \frac{4}{\kappa^2}} \nonumber \\
    &= \left( \frac{2}{\kappa^2} + 1 \right) - \frac{2}{\kappa^2}\sqrt{\kappa^2 + 1} \nonumber \\
    &= 1 - 2\frac{\sqrt{\kappa^2 + 1} - 1}{(\kappa^2 + 1) - 1} \nonumber \\
    &= 1 - \frac{2}{\sqrt{\kappa^2 + 1} + 1}
\end{align}
% The expansion for small $\kappa \ll 1$ is given by
% \begin{align}
%     \chi &= \left( \frac{2}{\kappa^2} + 1 \right) - \frac{2}{\kappa^2}\sqrt{1 + \kappa^2} \nonumber \\
%     &= \left( \frac{2}{\kappa^2} + 1 \right) - \frac{2}{\kappa^2} \left(1 + \frac{\kappa^2}{2} - \frac{\kappa^4}{8} + \mathcal{O}\left(\kappa^6\right)\right) \nonumber \\
%     &= \frac{\kappa^2}{4} + \mathcal{O} \left( \kappa^4 \right)
% \end{align}
% The expansion for large $\kappa \gg 1$ is given by
% \begin{align}
%     \chi &= \left( \frac{2}{\kappa^2} + 1 \right) - \frac{2}{\kappa} \left(1 + \frac{1}{2\kappa^2} - \frac{1}{8\kappa^4} + \mathcal{O}\left(\frac{1}{\kappa^6}\right)\right) \nonumber \\
%     &= 1 - \frac{2}{\kappa} + \mathcal{O}\left(\frac{1}{\kappa^2}\right)
% \end{align}
Going back to the recurrence, and given that the recurrence on $y_n$ can be shown to be the same as eq.~\ref{eq:ss2_clean_rec_x_n}, we get that if $(\bx^*, \by^*)$ is a stationary point of $f$ in $\ell_2 \times \ell_2$, then
\begin{align}
    \bx^*(i) &= x_{i} = c_1 \chi^i \\
    \by^*(i) &= y_{i} = c_2 \chi^i
\end{align}
where $c_1, c_2$ can be determined from the initial conditions given in eq.~\ref{eq:ss2_x_1} and \ref{eq:ss2_y_1}.
Using the domino argument, which yields that $\forall i > t+1, \bx_t(i) = 0$, we get that the distance to the optimum of $\bx$ is given by
\begin{align}\label{eq:using_domino_to_lower_bound}
    \norm{\bx_t-\bx^*}^2 = \sum_{i=1}^{t+1}(\bx_t(i)-\bx^*(i))^2 + \sum_{i=t+2}^{\infty} (\bx_t(i)-\bx^*(i))^2 &\geq \sum_{i=t+2}^{\infty} \left(\bx^*(i)\right)^2 \nonumber \\
    &= c_1^2 \sum_{i=t+2}^{\infty} \chi^{2i} \nonumber \\
    &= c_1^2 \sum_{i=1}^{\infty} \chi^{2\left(i+t+1\right)} \nonumber \\
    &= \chi^{2(t+1)} \norm{\bx^*}^2
\end{align}
% Therefore, asymptotically, the lower bound we get for very large $\kappa$ is
% \begin{align}
%     \norm{\bx_t-\bx^*}^2 &\gtrapprox \left(1 - \frac{2}{\kappa}\right)^{2t} \norm{\bx^*}^2
% \end{align}
Similarly, we can show that $\norm{\by_t-\by^*}^2 \geq \chi^{2(t+1)} \norm{\by^*}^2$.

Changing back our variables to $\bx \rightarrow \bx - \bx_0,\, \by \rightarrow \by - \by_0$ yields the bound for arbitrary initialisation.

\subsection{Proof of Thm. \ref{thm:domino_good_lower_bound_infdim}}\label{apdx:domino_good_lower_bound_infdim_proof}
If one computes $\mu_{12}$ for the function used in the previous bound, it becomes clear that $\mu_{12} = 0$. As such, it is not surprising that the bound failed to hold vs the upper bound of negative momentum on bilinear games with $\mu_{12} > 0$: the previous bound failed to be general enough because the example has the worst possible value of $\mu_{12}$. An important note, however, is that the previous bound may still hold in finite dimensions if we only used it to lower bound the rate of convergence of games with $\mu_{12} = 0$, but it can easily be checked that the rate in the improved bound with $\mu_{12} = 0$ reduces to the first bound.

In order to address this, we will pick values of $a_0$ and $a_1$ that allow the counterexample to handle any value of $\mu_{12}$. The proof of the improved domino bound follows the same line of argumentation as the proof of the first bound. We resume from eq. \ref{eq:apdx-domino-spectrum-checkpoint}, and set $c=1$, and suppose that $a_1 < 0, a_0 > 0$ such that $|a_1| \leq a_0$.  Theorem 7.20 of \citet{douglas2012banach} yields that:
\begin{align}
    \max \sigma(\mat{MM}^\top) &= a_0^2 + a_1^2 - 2a_0a_1 = (a_0 - a_1)^2\\
    \min \sigma(\mat{MM}^\top) &= a_0^2 + a_1^2 + 2a_0a_1 = (a_0 + a_1)^2
\end{align}
Thus, we have $\mu_{12}^2 = (a_0 + a_1)^2,\ L_{12}^2 = (a_0 - a_1)^2$ and since we assumed $|a_1| \leq a_0$, we get that $\mu_{12} = a_0 + a_1, L_{12} = a_0 - a_1$ which allows us to choose $a_0, a_1$ to make $\mu_{12}, L_{12}$ appear in the bound:
\begin{align}
    a_0 = \frac{L_{12} + \mu_{12}}{2} \\
    a_1 = \frac{\mu_{12} - L_{12}}{2}
\end{align}
Noting further that $a_0^2 + a_1^2 = \frac{L_{12}^2 + \mu_{12}^2}{2},\ a_0^2 - a_1^2 = \mu_{12}L_{12}$, we have that
\begin{align}
    \chi &= \frac{-\left( \frac{\mu_1 \mu_2}{c^2} + a_0^2 + a_1^2 \right)+ \sqrt{\left(\frac{\mu_1 \mu_2}{c^2}\right)^2 + 2 \frac{\mu_1 \mu_2}{c^2}\left(a_0^2 + a_1^2 \right) + \left(a_0^2 - a_1^2 \right)^2}}{2 a_0 a_1} \nonumber\\
    &= \frac{-\left( \mu_1 \mu_2 + \frac{L_{12}^2 + \mu_{12}^2}{2} \right)+ \sqrt{\left(\mu_1 \mu_2\right)^2 + 2 \mu_1 \mu_2\left(\frac{L_{12}^2 + \mu_{12}^2}{2} \right) + \left(\mu_{12}L_{12} \right)^2}}{\frac{\mu_{12}^2 - L_{12}^2}{2}} \nonumber\\
    &= \frac{-\left( 2 \mu_1 \mu_2 + L_{12}^2 + \mu_{12}^2 \right)+ 2\sqrt{\left(\mu_1 \mu_2\right)^2 + \mu_1 \mu_2\left(L_{12}^2 + \mu_{12}^2 \right) + \left(\mu_{12}L_{12} \right)^2}}{\mu_{12}^2 - L_{12}^2 + \mu_1\mu_2 - \mu_1 \mu_2}
\end{align}
Letting $d_{\mu} = \mu_1\mu_2 + \mu_{12}^2,\ d_{L} = \mu_1 \mu_2 + L_{12}^2$,
\begin{align}
    \chi &= \frac{-\left( d_{\mu} + d_L \right)+ 2\sqrt{\mu_1\mu_2\left(\mu_1 \mu_2 + L_{12}^2\right) + \mu_{12}^2\left(\mu_1\mu_2 + L_{12}^2 \right)}}{d_ {\mu} - d_L} \nonumber \\
    &= \frac{\left( d_{\mu} + d_L \right)- 2\sqrt{d_{\mu}d_L}}{d_L - d_{\mu}} \nonumber \\
    &= \frac{\left(\sqrt{d_L} - \sqrt{d_{\mu}}\right)^2}{\sqrt{d_L}^2 - \sqrt{d_{\mu}}^2} \nonumber \\
    &= \frac{\sqrt{d_L} - \sqrt{d_{\mu}}}{\sqrt{d_L} + \sqrt{d_{\mu}}} \nonumber\\
    &= 1 - \frac{2}{\sqrt{\frac{d_L}{d_{\mu}}} + 1}
\end{align}
Letting $\kappa = \frac{d_L}{d_{\mu}} = \frac{L_{12}^2 + \mu_1\mu_2}{\mu_{12}^2 + \mu_1\mu_2}$ and proceeding as in the proof of the previous bound with the new value of $\chi$ yields Thm. \ref{thm:domino_good_lower_bound_infdim}. Note that as promised, this rate reduces to that of the first bound if $\mu_{12} = 0$:
\begin{align}
    1 - \frac{2}{\sqrt{\frac{d_L}{d_{\mu}}} + 1} &= 1 - \frac{2}{\sqrt{\frac{L_{12}^2 + \mu_1\mu_2}{\mu_1\mu_2}} + 1} \nonumber\\
    &= 1 - \frac{2}{\sqrt{\kappa_{old}^2 + 1} + 1} \\
    % &= 1 - \frac{2\left(\sqrt{\kappa_{old}^2 + 1} - 1\right)}{\kappa_{old}^2} \nonumber\\
    % &= \frac{2}{\kappa_{old}^2} + 1 - \frac{2}{\kappa_{old}}\sqrt{1 + \frac{1}{\kappa_{old}^2}}
\end{align}
where $\kappa_{old}$ is the condition number of eq. \ref{eq:kappa_bad_bound}.

\newpage
\section{Proofs of $p$-SCLI-$n$}
\subsection{Proof of Prop. \ref{prop:spectral_lower_bound}}\label{apdx:proof_spectral_lower_bound}
In this section, we follow \citet{arjevani2016lower} to derive results for the $p$-SCLI-$n$ methods. First, we reproduce several definitions and theorems that are proven in \citet{arjevani2016lower} and that apply directly to the generalisation. Here, $\mat{A}$ will denote the Jacobian of some quadratic game with $f_{\mat{A},\bb} \in \cQ^{d_1,...,d_n}$ such that 0 is not in the spectrum of $\mat{A}$.

\begin{definition}[Characteristic polynomial of a $p$-SCLI-$n$] \label{def:char_poly}
Let $\cA$ be a $p$-SCLI-$n$ optimisation algorithm with coefficient matrices $C_i$ as defined in def. \ref{def:p-SCLI-n}. Then for $\mat{X} \in \reals^{d\times d}$, the characteristic polynomial of $\cA$ is given by
\begin{align} 
	\cL (\lambda,\mat{X}) &\triangleq  \mat{I}_d\lambda^p - \sum_{i=0}^{p-1} \EV C_i(\mat{X}) \lambda^{i}  
\end{align}
and its root radius is
\begin{align*}
	\rho_\lambda(\cL (\lambda,\mat{X}))&= \rho(\det{\cL(\lambda,\mat{X})}) = \max\left\{\absval{\lambda} \ \middle\vert \ \det{\cL(\lambda,\mat{X})}=0\right\} 
\end{align*}
\end{definition}

\begin{thm} [Consistency - characteristic polynomial (Based on Theorem 5 of \citet{arjevani2016lower})]\label{thm:consistency-char_poly}
A $p$-SCLI-$n$ algorithm $\cA$ with characteristic polynomial $\cL(\lambda, \mat{X})$ and inversion matrix $N(\mat{X})$ is consistent with respect to $\mat{A}$ if and only if the following two conditions hold:
\begin{align}
&1.\ \cL (1,\mat{A}) = -\EV N(\mat{A})\mat{A} \label{eq:char_poly_1_invmat}\\
&2.\ \rho_\lambda(\cL (\lambda,\mat{A})) < 1 \label{eq:root_radius_char_poly}
\end{align}
\end{thm}

We may rephrase theorem 13 of \citet{arjevani2016lower} (and lower bound $t^{m-1}$ by $1$ since $m \in \mathbb{N}$)  as the following to use the root radius of the characteristic polynomial to show linear rates:
\begin{thm}[Based on Theorem 13 of \citet{arjevani2016lower}]\label{thm:lower_bound_root_radius}
If $\mat{A}$ is the Jacobian of a quadratic game and $\cA$ is a $p$-SCLI-$n$, there exists an initialisation point $\bw_0 \in \reals^d$ such that 
\begin{align}
    \max_{i=0,\dots,p-1} \norm{\EV\bw^{t+i} - \EV\bw^*} \in \Omega(\rho_\lambda(\cL (\lambda,\mat{A}))^t)
\end{align}
\end{thm}
In other words, this means that $\cA$ cannot converge on $f_{\mat{A}, \bb}$ with linear rate faster than $\rho_\lambda(\cL (\lambda,\mat{A}))$, up to a constant. As \citet{arjevani2016lower} argue, in both deterministic and stochastic settings, a lower bound on $\norm{\EV \left[\bw^t - \bw^* \right]}^2$ implies\footnote{Note that since we only use in this paper a lower bound on the second term of the right hand-side of the equation to bound the left hand-side, one may derive in stochastic settings tighter lower bounds than the ones presented in this paper by factoring in the first term of the right hand-side. We leave this as future work.} a lower bound on  $\EV \norm{\bw^t - \bw^*}^2$, since
\begin{align}
    \EV \left[\norm{\bw^t - \bw^*}^2\right] = \EV \left[\norm{\bw^t - \EV \bw^t}^2\right] + \norm{\EV \left[\bw^t - \bw^* \right]}^2 
\end{align}
We can now focus on finding a lower bound on $\rho_\lambda(\cL (\lambda,\mat{A}))$.
\begin{prop}\label{prop:rho_ineq_char_poly}
Let $\cA$ be a $p$-SCLI-$n$ optimisation algorithm with inversion matrix $N(\mat{X})$ that is consistent with respect to $\mat{A}$. Then,  
\begin{align} 
	\rho_\lambda(\cL (\lambda,\mat{A}))&\ge \max_{j=1,...,d} \absval{\sqrt[p]{|\sigma_j(-\EV[N(\mat{A})]\mat{A})|}-1} 
\end{align}
where the $\sigma_j(-\EV[N(\mat{A})]\mat{A})$ are elements of the spectrum (eigenvalues) of $-\EV[N(\mat{A})]\mat{A}$.
\end{prop}

\subsubsection{Proof of Prop. \ref{prop:rho_ineq_char_poly}}\label{apdx:proof_rho_ineq_char_poly}
Our proof starts exactly as the one presented by \citet{arjevani2016lower} for the $n=1$ particular case, where the authors assume that $\mat{A}$ is symmetric with strictly positive spectrum. However, we will generalise the proof to cover non-symmetric matrices and matrices that may not have strictly positive spectrum, since the Jacobian of a quadratic $n$-player game generally does not have these properties. 

Let $\cA$ be a deterministic $p$-SCLI-$n$ optimisation algorithm with characteristic polynomial $\cL(\lambda,\mat{X})$ and inversion matrix $N(\mat{X})$, and $f_{A,\bb}(\bw)\in\cQ^{d_1,...,d_n}$ represent a quadratic $n$-player game.
Since $\cA$ is $p$-SCLI-$n$, its (expected) coefficient matrices $\EV C_i$ evaluated on $\mat{A}$ are simultaneously triangularisable, so $\exists \mat{Q}\in\reals^{d\times d}$ such that for $i= 0, ..., p-1$, we have
\begin{align}
    \mat{T}_i\triangleq \mat{Q}^{-1} \EV C_i(\mat{A}) \mat{Q}
\end{align}
where $\mat{T}_i$ is triangular. Thus,
\begin{align}
		\det \cL (\lambda,\mat{A})  &= \det\lrpar{\mat{Q}^{-1} \cL (\lambda,\mat{A}) \mat{Q} } \nonumber \\
		&= \det\lrpar{\mat{I}_{d}\lambda^p - \sum_{i=0}^{p-1} \mat{T}_i\lambda^{i}}
\end{align}
Since $\mat{I}_{d}\lambda^p - \sum_{i=0}^{p-1} \mat{T}_i\lambda^{i}$ is a upper triangular matrix, its determinant is given by
\begin{align}
    \det \cL (\lambda,\mat{A}) = \prod_{j=1}^{d} \ell_j(\lambda)
\end{align}
where
\begin{align}
    \ell_j(\lambda) &= \lambda^p - \sum_{i=0}^{p-1} \sigma_j^i\lambda^{i} 
\end{align}
and where $\sigma_1^i,\dots, \sigma_{d}^{i},\ i=0,\dots,p-1$ denote the elements on the diagonal of $\mat{T}_i$, which are just the eigenvalues of $\EV C_i$ ordered according to $\mat{Q}$. Hence, the root radius of the characteristic polynomial of $\cA$ is
\begin{align} \label{eq:max_over_ells}
	\rho_\lambda(\cL (\lambda,\mat{A})) &=  \max \left\{\absval{\lambda} \ \middle\vert \ \ell_j(\lambda)=0 \text{ for some } j=1,...,d\right\} 
\end{align}
On the other hand, by consistency condition (\ref{eq:char_poly_1_invmat}) we get that for all $j=1,...,d$ 
\begin{align}
	\ell_j(1)=\sigma_j\lrpar{\cL (1, \mat{A})}=\sigma_j\lrpar{-\EV[N(\mat{A})]\mat{A}} 
\end{align}

In the case of $p$-SCLI-1, the authors prove their Corollary 7 (i.e. our prop. \ref{prop:rho_ineq_char_poly} without taking the modulus of the eigenvalues) by using a lemma (see Lemma 6 in \citet{arjevani2016lower}) that gives a lower bound on each $\rho(\ell_j(\lambda))$ by using the sign of $\ell_j(1)=\sigma_j\lrpar{-\EV[N(\mat{A})]\mat{A}}$. Lemma 6 of \citet{arjevani2016lower} is proven using the following lemma, which we can in fact use to handle arbitrary eigenvalues (e.g. complex or negative).
\begin{lemma}[Lemma 15 of \citet{arjevani2016lower}] \label{lem:eco_poly}
Let $q_r^*(z) \triangleq \lrpar{z-(1-\sqrt[p]{r})}^p$ where $r$ is some non-negative constant. Suppose $q(z)$ is a monic polynomial of degree $p$ with complex coefficients. Then,
\begin{align*}
\rho(q(z))\le \absval{\sqrt[p]{\absval{q(1)}}-1} \iff q(z)=q_{\absval{q(1)}}^*(z)
\end{align*}
\end{lemma}
The proof of the lemma can be found in \citet{arjevani2016lower}. Here, we can use the lemma directly on each $\ell_j$ with $q = \ell_j$ and $r = |q(1)| = |\ell_j(1)| = |\sigma_j\lrpar{-\EV[N(\mat{A})]A}|$. Indeed, since $r \geq 0$, 
\begin{itemize}
    \item if $q(z) = q_r^*(z) = (z - (1 - \sqrt[p]{r}))^p$ then clearly $\rho(q(z)) = |1 - \sqrt[p]{r}|$
    \item if $q(z) \neq q_r^*(z)$, then we have $\rho(q(z)) > |\sqrt[p]{|q(1)|} - 1|$
\end{itemize}
Which implies that for any $j$ we have $\rho(\ell_j(\lambda)) \geq |\sqrt[p]{|\ell_j(1)|} - 1| = |\sqrt[p]{|\sigma_j\lrpar{-\EV[N(\mat{A})]\mat{A}}|} - 1|$. Using this in eq. \ref{eq:max_over_ells} yields
\begin{align}\label{eq:lower_bound_on_root_radius}
	\rho_\lambda(\cL (\lambda,\mat{A}))&\ge \max_{j=1,...,d} \absval{\sqrt[p]{|\sigma_j(-\EV[N(\mat{A})]\mat{A})|}-1} 
\end{align}

\subsubsection{Deriving the optimal $\rho$ for scalar inversion matrices}
We are now ready to obtain the general lower bound. Consider $f_{\mat{A}, \bb}\in \cQ^{d_1,...,d_n}$ with $0 \not \in \sigma(\mat{A})$ and a consistent $p$-SCLI-$n$ algorithm $\cA$. Let $\mu = \min |\sigma(\mat{A})|$, $L = \max |\sigma(\mat{A})|$ where $\sigma(\mat{A})$ is the spectrum of $\mat{A}$. For a scalar inversion matrix i.e. $\EV[N(\mat{A})] = \nu$ we have from eq. \ref{eq:lower_bound_on_root_radius}:
\begin{align}
	\rho_\lambda(\cL (\lambda,\mat{A}))&\ge \max_{j=1,...,d} \absval{\sqrt[p]{|\sigma_j(-\EV[N(\mat{A})]\mat{A})|}-1} 
	= \max_{j=1,...,d} \absval{\sqrt[p]{|\nu\sigma_j(\mat{A})|}-1} \nonumber \\
	&= \max\left\{|\sqrt[p]{|\nu| \mu}-1|, |\sqrt[p]{|\nu| L}-1| \right\}	
\end{align}
Note that consistency (eq. \ref{eq:root_radius_char_poly}) constrains $\nu \in \left(\frac{-2^p}{L}, \frac{2^p}{L}\right)\setminus\{0\}$. We proceed as \citet{arjevani2016lower} in the $p$-SCLI-$1$ case, and study the ranges of $|\nu|$ by using $\max(a, b) = \frac{a+b + |a-b|}{2}$ to obtain table \ref{table:nu_cases}.
\begin{table}[H] 
		\caption{Lower bound for $\rho$ by subranges of $|\nu|$ and minimiser $|\nu^*|$} \label{table:nu_cases}
		\centering
    \begin{tabularx}{0.9\textwidth}{@{}lYYYYYY@{}} 
    \toprule
        &\multicolumn{3}{c}{$\sqrt[p]{|\nu| \mu}-1 < 0$} & \multicolumn{3}{c}{$\sqrt[p]{|\nu| \mu}-1 \ge 0$} \\
        \cmidrule(lr){2-4} \cmidrule(l){5-7} & Range & Minimiser & Bound & Range & Minimiser & Bound \\
		\midrule
		$\sqrt[p]{|\nu| L}-1 \le 0$  & $(0,1/L]$ & $1/L$ & $1-\sqrt[p]{\frac{\mu}{L}}$ & & N/A & \\
		&&&&&&\\
		$\sqrt[p]{|\nu| L}-1 > 0$ & $(1/L,1/\mu)$ & $\lrpar{\frac2{\sqrt[p]{ L} +\sqrt[p]{\mu} }}^p$ & $\frac{\sqrt[p]{ L/\mu} -1 }{\sqrt[p]{L/\mu} +1}$ 
		& $[1/\mu, 2^p/L)$ & $1/\mu$ & $\sqrt[p]{\frac{L}{\mu}}-1$ \\
		\bottomrule
		\end{tabularx}
\end{table}
Note that case 3 requires $p> \log_2L/\mu$.
Hence,
\begin{align}
\rho \ge \min \left\{ 1-\sqrt[p]{\frac{\mu}{L}},\frac{\sqrt[p]{ L/\mu} -1 }{\sqrt[p]{L/\mu} +1}, \sqrt[p]{\frac{L}{\mu}}-1 \right\} 
= \frac{\sqrt[p]{L/\mu}-1}{\sqrt[p]{L/\mu}+1}
\end{align}
where $\mu = \min |\sigma(\mat{A})|$, $L = \max |\sigma(\mat{A})|$.

\subsection{Finding a suitably hard example for 2-player with $d_1=d_2$} \label{apdx:p-SCLI-2-refined-lower}
We now only need to find a hard counterexample.
We present the argument for $d_1 = d_2 = 2$, which can easily be generalised for arbitrary $d$. Consider the matrix
\begin{align}
    \mat{A} = \begin{pmatrix}
        \mu_1 &     0 & \mu_{12} & 0 \\
        0     & L_{1} & 0        & L_{12} \\
        -\mu_{12}& 0   & \mu_2    & 0 \\
        0       & -L_{12}& 0     & L_{2}
    \end{pmatrix}
\end{align}
corresponding to the Jacobian of a quadratic game in $\cQ^{d_1, d_2}$.

First we compute the characteristic polynomial of $A$, using the formula for the determinant of a block matrix (see \citet[Section 0.3]{zhangSchurComplementIts2005} for instance):
\begin{align}
\det(XI - A) &= \det\begin{pmatrix}
        X - \mu_1 &     0 & -\mu_{12} & 0 \\
        0     & X - L_{1} & 0        & -L_{12} \\
        \mu_{12}& 0   & X -\mu_2    & 0 \\
        0       & L_{12}& 0     & X - L_{2}
    \end{pmatrix}\\
    &= \det\left(\begin{pmatrix}
        (X - \mu_1)(X - \mu_2) &     0  \\
        0     & (X - L_{1})(X - L_{2})  \end{pmatrix}
        + \begin{pmatrix}
        \mu_{12}^2& 0   \\
        0       & L_{12}^2
    \end{pmatrix}
    \right)\\
    &= (X^2 - (\mu_1 + \mu_2)X + \mu_1\mu_2 + \mu_{12}^2)(X^2 - (L_1 + L_2)X + L_1L_2 + L_{12}^2)
\end{align}
 The discriminants of these two quadratic equations are, respectively:
 \begin{align}
&\Delta_\mu = (\mu_1 + \mu_2)^2 - 4(\mu_1\mu_2 + \mu_{12}^2) = (\mu_1 - \mu_2)^2 - 4 \mu_{12}^2\\
&\Delta_L   = (L_1 + L_2)^2 - 4(L_1L_2 + L_{12}^2) = (L_1 - L_2)^2 - 4 L_{12}^2
 \end{align}
 
which yields the following eigenvalues:
\begin{align}
    \lambda_{\mu\pm} &= \frac{\mu_1 + \mu_2}{2} \pm \sqrt{\left(\frac{\mu_1 - \mu_2}{2}\right)^2-\mu_{12}^2} \nonumber\\
    \lambda_{L\pm} &= \frac{L_1 + L_2}{2} \pm \sqrt{\left(\frac{L_1 - L_2}{2}\right)^2-L_{12}^2}
\end{align}
We distinguish four cases, which are presented in the following table:

\begin{table}[H]
\centering
\caption{Lower bounds on the condition number}
\begin{tabularx}{0.43\textwidth}{@{}lcc@{}} 
\toprule
                      & $\Delta_\mu < 0$                 & $\Delta_\mu \geq 0$                                                         \\
\midrule
$\Delta_L < 0$ & $\kappa = \sqrt{\frac{L_1L_2 + L_{12}^2}{\mu_1\mu_2 + \mu_{12}^2}}$ & $\kappa \geq 2\frac{\sqrt{L_1L_2 + L_{12}^2}}{\mu_1+\mu_2 - \sqrt\Delta_\mu}$                          \\ 
$\Delta_L \geq 0$         & $\kappa \geq \frac{1}{2}\frac{L_1+L_2 + \sqrt\Delta_L}{\sqrt{\mu_1\mu_2 + \mu_{12}^2}}$& $\kappa \geq \frac{L_1+L_2 + \sqrt\Delta_L}{\mu_1+\mu_2 - \sqrt\Delta_\mu}$ \\
\bottomrule
\end{tabularx}
\end{table}

where we used that $\kappa = \frac{\max{|\sigma(\mat{A})|}}{\min{|\sigma(\mat{A})|}}$. 

We now discuss these four cases:
\begin{itemize}
    \item If $\Delta_\mu < 0$ and $\Delta_L < 0$, we have that
    \begin{align}
    |\lambda_{\mu\pm}| &= \left|\frac{\mu_1 + \mu_2}{2} \pm i\sqrt{\mu_{12}^2 - \left(\frac{\mu_1 - \mu_2}{2}\right)^2}\right| \nonumber\\
    &= \sqrt{\mu_1\mu_2 + \mu_{12}^2}
\end{align}
Similarly we get
\begin{align}
    |\lambda_{L\pm}| &= \sqrt{L_1 L_2 + L_{12}^2}
\end{align}
Clearly then $\min{|\sigma(\mat{A})|} = |\lambda_{\mu\pm}|$ and $\max{|\sigma(\mat{A})|} = |\lambda_{L\pm}|$, which yields $\kappa = \sqrt{\frac{L_1L_2 + L_{12}^2}{\mu_1\mu_2 + \mu_{12}^2}}$. 
\item If $\Delta_\mu \geq 0$ and $\Delta_L \geq 0$, $\lambda_{L+},\ \lambda_{L-},\ \lambda_{\mu+}$ and $\lambda_{\mu-}$ are all real. We have that,
\begin{align}
\lambda_{\mu-} \geq \min |\sigma (\mat{A})|\,,\quad\text{and}\quad  \lambda_{L+} \leq \max |\sigma(\mat{A})|\,,
\end{align}
which yields the result.
\item If $\Delta_\mu < 0$ and $\Delta_L \geq 0$, it holds that,
\begin{align}
|\lambda_{\mu\pm}| = \min |\sigma (\mat{A})|\,,\quad\text{and}\quad  \lambda_{L+} \leq \max |\sigma(\mat{A})|\,,
\end{align}
from which we obtain the result.
\item Similarly, if $\Delta_\mu \geq 0$ and $\Delta_L < 0$, it holds that,
\begin{align}
\lambda_{\mu-} \geq \min|\sigma (\mat{A})|\,,\quad\text{and}\quad  |\lambda_{L\pm}| = \max |\sigma(\mat{A})|\,.
\end{align}
\end{itemize}
One could wonder whether our lower bounds on $\kappa$ when at least one of the discriminant is non-negative are actually equalities. We provide an example showing that it is not the case when $\Delta_L \geq 0$ and $\Delta \mu \geq 0$. A similar one can be found when   $\Delta_L < 0$ and $\Delta \mu \geq 0$.

Take $\mu_{12}=0$ and $L_{12} = \frac{|L_1 - L_2|}{2}$. Then $\Delta_L \geq 0$ and $\Delta \mu \geq 0$. Then,
\begin{align}
\lambda_{\mu+} &= \frac{\mu_1 + \mu_2}{2} + \sqrt{\left(\frac{\mu_1 - \mu_2}{2}\right)^2-\mu_{12}^2} = \max(\mu_1, \mu_2)\nonumber\\
\lambda_{L\pm} &= \frac{L_1 + L_2}{2} \pm \sqrt{\left(\frac{L_1 - L_2}{2}\right)^2-L_{12}^2} = \frac{L_1 + L_2}{2}\,.
\end{align}
Choose $\mu_1 = L_1$, $\mu_2 = L_2$ and $L_1 \neq L_2$. Then $\lambda_{\mu+} > \lambda_{L\pm}$. However we have $\lambda_{\mu-} = \min |\sigma(A)|$ and so in this case $\kappa = \lambda_{\mu+}/\lambda_{\mu-}$.

\newpage
\section{Lower bounds on the iteration complexity}\label{apdx:iteration-complexity}
The lower bounds on the distance of the iterates to a minimiser also yield lower bounds on the number of iterations required to reach a maximum error $\epsilon$ on the iterates (iteration complexity). 

\textbf{$p$-SCLI-$n$ case:} For the $p$-SCLI-$n$ bound, suppose $\max_{i=0,\dots,p-1} \norm{\EV\bw^{t+i} - \EV\bw^*} < \epsilon$. Then from Theorem \ref{thm:lower_bound_root_radius}, we know that for some strictly positive real $C$,
\begin{align}
    C \rho_\lambda(\cL (\lambda,\mat{A}))^t \leq \epsilon &\implies \rho_\lambda(\cL (\lambda,\mat{A}))^t \leq \frac{\epsilon}{C} \nonumber \\
    &\implies t \log\lrpar{\rho_\lambda(\cL (\lambda,\mat{A}))} \leq \log\lrpar{\frac{\epsilon}{C}} \nonumber \\
    &\implies t \log\lrpar{\frac{1}{\rho_\lambda(\cL (\lambda,\mat{A}))}} \geq \log\lrpar{\frac{C}{\epsilon}}
\end{align}
Noting that $\log(x) \leq x - 1$ for $x > 0$, we get that $\log\lrpar{\frac{1}{\rho_\lambda(\cL (\lambda,\mat{A}))}} \leq \frac{1 - \rho_\lambda(\cL (\lambda,\mat{A}))}{\rho_\lambda(\cL (\lambda,\mat{A}))}$ and hence,
\begin{align}
    t \frac{1 - \rho_\lambda(\cL (\lambda,\mat{A}))}{\rho_\lambda(\cL (\lambda,\mat{A}))} \geq t \log\lrpar{\frac{1}{\rho_\lambda(\cL (\lambda,\mat{A}))}} \geq \log\lrpar{\frac{C}{\epsilon}}
\end{align}
Therefore, we get that
\begin{align}
    t \geq \frac{\rho_\lambda(\cL (\lambda,\mat{A}))}{1 - \rho_\lambda(\cL (\lambda,\mat{A}))} \log\lrpar{\frac{C}{\epsilon}}
\end{align}
where one may use Prop. \ref{prop:spectral_lower_bound} to get
\begin{align}
    t \geq \lrpar{\frac{\sqrt[p]{\kappa}-1}{2}} \log\lrpar{\frac{C}{\epsilon}}
\end{align}
where $\kappa$ is given as in Prop. \ref{apdx:proof_spectral_lower_bound}. One may also use in the 2-player case the $\kappa$ given in Table \ref{table:p-SCLI-2_kappa_lower_bounds}.

\textbf{2-player domino bound:} For the two player bound stemming from the domino argument, it is equally straightforward to establish from Thm. \ref{thm:domino_good_lower_bound_infdim} that given an upper bound $\epsilon$ on $\norm{(\bx_t, \by_t)-(\bx^*, \by^*)}$, then using that $1-\frac{2}{\kappa+1} \geq \exp{\lrpar{-\frac{2}{\kappa-1}}}$,
\begin{align}
    \exp{\lrpar{-\frac{2}{\kappa-1}}}^{t+1} \cdot \norm{(\bx_0, \by_0) - (\bx^*, \by^*)} \leq \epsilon \implies t \geq \frac{\kappa-1}{2}\log\lrpar{\frac{\norm{(\bx_0, \by_0) - (\bx^*, \by^*)}}{\epsilon}} - 1
\end{align}
where $\kappa$ is defined as in Thm. \ref{thm:domino_good_lower_bound_infdim}.
\end{appendices}
\end{document}